\pdfoutput=1

\documentclass[11pt]{article}

\usepackage[final]{acl}
\usepackage{hyperref}

\usepackage{times}
\usepackage{latexsym}

\usepackage[T1]{fontenc}

\usepackage[utf8]{inputenc}

\usepackage{microtype}

\usepackage{inconsolata}
\usepackage{multirow}
\usepackage{framed}
\usepackage{makecell} 
\usepackage{lipsum} 
\usepackage{enumitem}
\usepackage{listings}
\usepackage{paralist}
\usepackage{booktabs}
\usepackage{graphicx}
\usepackage{amssymb}
\usepackage{pifont}
\usepackage{marginnote}
\usepackage[color=white,size=footnotesize]{todonotes}
\usepackage{verbatim}

\title{Commonsense Knowledge with Negation: \\A Resource to Enhance Negation Understanding}

 \author{Zijie Wang\textsuperscript{1} \quad
         MohammadHossein Rezaei\textsuperscript{1} \quad
         Farzana Rashid\textsuperscript{2} \quad
         Eduardo Blanco\textsuperscript{1}\\
         \textsuperscript{1}University of Arizona \quad
         \textsuperscript{2}University of North Carolina Asheville\\
         \texttt{\{zijiewang, mhrezaei, eduardoblanco\}@arizona.edu} \quad \texttt{frashid@unca.edu}
 }

\begin{document}
\maketitle
\begin{abstract}
  Negation is a common and important semantic feature in natural language,
  yet Large Language Models (LLMs) struggle when negation is involved in natural language understanding tasks.
  Commonsense knowledge, on the other hand, despite being a well-studied topic,
  lacks investigations involving negation.
  In this work, we show that commonsense knowledge with negation is challenging for models to understand.
  We present a novel approach to automatically augment existing
  commonsense knowledge corpora with negation, yielding two new corpora containing over 2M triples with \emph{if-then} relations.
  In addition, pre-training LLMs on our corpora benefits negation understanding.
\end{abstract}

\section{Introduction}

Negation is a common and important semantic feature in natural language,
appearing in approximately 25\% of English sentences~\cite{hossain-etal-2020-analysis}.
Despite the recent success of large language models (LLMs) across various natural language processing tasks~\cite{achiam2023gpt,touvron2023llama},
their understanding of negation remains unclear.
Previous work has demonstrated that language models struggle with
multiple natural language understanding tasks when negation is involved~\cite{dobreva-keller-2021-investigating, jang-etal-2022-beyond}.
However, these investigations have been limited to encoder-based models such as BERT~\cite{devlin-etal-2019-bert} and earlier LLMs such as GPT-3~\cite{brown2020language}.

Commonsense knowledge has been extensively studied,
with numerous efforts focused on building commonsense knowledge bases~\cite{speer2017conceptnet,sap2019atomic}.
Commonsense reasoning has also been investigated through tasks such as question answering~\cite{talmor-etal-2019-commonsenseqa}.
Despite these extensive efforts,
the intersection of commonsense knowledge and negation remains underexplored.

\begin{figure}[t!]
  \centering
  \small
  \includegraphics[width=\columnwidth]{figure/intro_figure-crop.pdf}
  \caption{
    A commonsense knowledge triple with the \emph{Intention} relation.
    Negations are added to both \emph{if} and \emph{then} events. 
    Adding different negation cues results in new triples that align (same color on both sides) or conflict with (different colors) commonsense knowledge.}
    \label{f:motivation}
\end{figure}

\textsc{Atomic}~\cite{sap2019atomic} is one of the largest commonsense knowledge corpora with \emph{if-then} relations,
representing commonsense knowledge as a triple <\textit{A}, \textit{R}, \textit{B}>,
where \textit{A} represents the \emph{if} event, \textit{R} represents the relation, and \textit{B} represents the \emph{then} event.
For example, <\emph{the person is hungry}, \emph{the person wants to}, \emph{eat food}> represents the commonsense knowledge that
``\emph{if the person is hungry, then the person wants to eat food}.''
To our knowledge, \textsc{Anion}~\cite{jiang-etal-2021-im}
is the only work that develops a commonsense knowledge base with negation,
built on top of \textsc{Atomic}.
However, their approach only negates the \emph{if} event and generates a new \emph{then} event via human annotation,
resulting in a new triple <\textit{$\neg$A}, \textit{R}, \textit{B$^\prime$}>.
This is limited because it does not consider negating the \emph{then} event
and requires significant human annotation effort.

In this work,
we present a novel approach to automatically augment existing commonsense knowledge corpora
with negation.
Our method is motivated by the observation that
negating either the \emph{if} event, the \emph{then} event, or both
can sometimes produce new triples that still align with commonsense knowledge.
As shown in the example in Figure~\ref{f:motivation}, negating the \emph{if} event
\emph{``Euler \textbf{unsuccessfully} applied for a position in Physics''}
implies that \emph{the intention of Euler} was to \emph{``teach in Physics.''}
In contrast, negating the \emph{then} event
\emph{``\textbf{not} teach in Physics''}
yields a triple that conflicts with commonsense knowledge (green \emph{if} event and red \emph{then} event).
This observation motivates us to augment both \emph{if} and \emph{then} events with negation, 
expanding existing commonsense knowledge corpora by up to three times (negating the \emph{if} event, the \emph{then} event, or both).
In addition, our approach avoids relying on human effort to generate new events such as \emph{``\textbf{not} join Math department''}.
We further propose an automatic validation method to verify
whether the new commonsense triples with negation align with commonsense knowledge.
More importantly, 
we demonstrate that they benefit LLMs' understanding of negation in general-purpose tasks.
The main contributions of this paper are:\footnote{Code and dataset available at \url{https://github.com/wang-zijie/commonsense_with_negation}}
\begin{compactitem}
    \item A novel approach to develop two commonsense knowledge corpora with over 2M triples containing negation.
    \item An automatic method to validate commonsense triples with negation.
    \item Evaluation on multiple models across five benchmarks, demonstrating that our corpora improve LLMs' understanding of negation.
  \end{compactitem}

\section{Related Work}
\label{s:related_work}

\paragraph{Commonsense knowledge bases} (CSKB) have been developed across multiple works with different focuses.
ConceptNet~\cite{speer2017conceptnet} represents taxonomic commonsense knowledge as a graph,
where each concept (i.e., word or phrase) is connected by a relation. For example,
\emph{a net} is used for \emph{catching fish}.
\citet{sap2019atomic} focus on relations between events rather than concepts,
proposing nine \emph{if-then} relations representing inferential knowledge.
For example, if \emph{X pays Y a compliment}, then Y wants to \emph{return
the compliment}. To our knowledge, \textsc{Anion}~\cite{jiang-etal-2021-im}
is the only work that investigates commonsense knowledge with negation.
It is built by negating the \emph{if} event from \textsc{Atomic} and manually annotating a new \emph{then} event.
For example, if \emph{X does not pay Y a compliment}, then the effect on Y is \emph{upset}.
This approach limits the possibility of negating the \emph{then} event and requires significant human annotation effort.
\citet{10.1145/3511808.3557484} develop a method to identify informative negations of commonsense concepts. 
The dataset complements existing CSKB as they often do not capture any negative concepts such as ``gorillas are not territorial.''
The UNcommonsense dataset~\cite{zhao-etal-2024-uncommonsense} collects data
involving unusual, unexpected, and unlikely situations, namely uncommonsense knowledge.
The primary task asks LLMs to generate reasonable explanations given contexts with uncommon outcomes.
Beyond manual efforts to create CSKB,
\citet{bosselut-etal-2019-comet} propose COMET, a system that trains a transformer model to automatically
generate the \emph{then} event given the \emph{if} event and relation.
\citet{west-etal-2022-symbolic} distill commonsense knowledge from LLMs to train a transformer for commonsense graph generation.
Unlike existing works, we propose a novel approach to incorporate negation into existing CSKB.
Our approach generates large-scale commonsense knowledge triples with negation while requiring no human effort.

\paragraph{Commonsense Knowledge for Downstream Tasks}
Commonsense knowledge has been shown to benefit many downstream tasks.
\citet{talmor-etal-2019-commonsenseqa} present a question answering dataset called CommonsenseQA that involves reasoning over commonsense knowledge.
They develop multiple-choice questions based on the concept-relation graph from ConceptNet.
\citet{lal-etal-2022-using} observe that answering why-questions often requires commonsense reasoning
and leverage COMET to generate relevant commonsense knowledge for this task.
\citet{guan-etal-2020-knowledge} pre-train a Transformer model on external
commonsense knowledge bases to improve story generation.
In this work, we leverage augmented commonsense knowledge corpora with negation to
improve LLMs' understanding of negation, with experimental results across three tasks demonstrating its effectiveness.

\paragraph{Improving Models' Negation Understanding}
Prior work has shown that LLMs struggle with 
natural language understanding tasks involving negation~\cite{dobreva-keller-2021-investigating, jang-etal-2022-beyond},
motivating efforts to address this limitation.
\citet{hosseini-etal-2021-understanding} leverage unlikelihood
training and synthetic data generation to improve BERT's understanding of negation.
\citet{singh-etal-2023-nlms} modify BERT's next sentence prediction task by incorporating negation cues rather than random sentences.
\citet{rezaei-blanco-2024-paraphrasing} show that
incorporating affirmative interpretations improves performance on negation understanding benchmarks.
In this work, we demonstrate that state-of-the-art LLMs still lack robust negation understanding 
and develop two commonsense knowledge corpora augmented with negation. 
More importantly, pre-training on our corpora improves performance on multiple tasks requiring negation understanding.

\begin{table*}[t!]
  \centering
  \small
\begin{tabular}{l rrr rrr rrr rrrr}
    \toprule
    & \multicolumn{3}{c}{Valid} & \multicolumn{3}{c}{Invalid} & \multicolumn{3}{c}{Ambiguous} & \multicolumn{4}{c}{Overall} \\
    \cmidrule(lr){2-4} \cmidrule(lr){5-7} \cmidrule(lr){8-10} \cmidrule(lr){11-14}
    & P & R & F1 & P & R & F1 & P & R & F1 & P & R & F1 & Acc \\
    \midrule
    Few-shot learning \\
    ~~~Llama 3.1 8B      & 0.56 & 0.58 & 0.57 & 0.40 & 0.71 & 0.52 & 0.54 & 0.07 & 0.13 & 0.45 & 0.45 & 0.37 & 0.44 \\
    ~~~Llama 3.1 70B     & 0.73 & 0.54 & 0.62 & 0.50 & 0.73 & 0.59 & 0.48 & 0.38 & 0.43 & 0.56 & 0.55 & 0.53 & 0.53 \\
    ~~~GPT-4o            & 0.71 & 0.48 & 0.51 & 0.54 & 0.65 & 0.59 & 0.51 & 0.54 & 0.51 & 0.53 & 0.54 & 0.52 & 0.54 \\
    ~~~Claude Sonnet 4   & 0.83 & 0.38 & 0.53 & 0.51 & 0.63 & 0.56 & 0.46 & 0.64 & 0.54 & 0.61 & 0.56 & 0.56 & 0.56 \\
    \addlinespace
    Fine-tuning \\
    ~~~Llama 3.1 8B      & 0.57 & 0.80 & 0.67 & 0.67 & 0.48 & 0.56 & 0.53 & 0.45 & 0.48 & 0.58 & 0.57 & 0.55 & 0.56 \\
    ~~~Llama 3.1 70B     & \textbf{0.70} & 0.76 & 0.73 & \textbf{0.79} & 0.48 & 0.59 & 0.51 & 0.68 & 0.58 & 0.65 & 0.63 & 0.63 & 0.64 \\
    \bottomrule
\end{tabular}

  \caption{
      Results of validating commonsense triples with negation via (1) few-shot learning with Llama 3.1 8B, 70B, GPT-4o, and Claude Sonnet 4,
      and (2) fine-tuning Llama 3.1 8B and 70B.
      We report Precision (P), Recall (R), and F1 for each label and overall, along with overall Accuracy (Acc).
      Only the best fine-tuning results are reported across variants of whether \textit{Valid} and \textit{Ambiguous} training instances come from \textsc{Atomic}, \textsc{Anion}, or both;
      \textit{Invalid} training data is synthesized using GPT-4o. Full results are reported in Appendix~\ref{app:additional_validating_results}.
    }
  \label{t:validation_results}

\end{table*}
\section{Augmenting Commonsense Knowledge Triples with Negation}
We propose a novel approach to automatically augment existing
commonsense knowledge corpora with negation.
Building on \textsc{Atomic}~\cite{sap2019atomic} and \textsc{Anion}~\cite{jiang-etal-2021-im},
we develop two new commonsense knowledge corpora:
$\neg$\textsc{Atomic} and $\neg$\textsc{Anion}.
We further introduce an automatic validation method that categorizes
the generated commonsense triples as either
\emph{Valid} (aligned with commonsense knowledge),
\emph{Invalid} (conflicting with commonsense knowledge), 
or \emph{Ambiguous} (neither).
Our resulting corpora contain over 2M commonsense triples with negation.

\noindent
\paragraph{Negating Commonsense Knowledge Triples}
As discussed in Section~\ref{s:related_work},
\textsc{Atomic} consists of commonsense knowledge triples with \emph{if-then} relations connecting two events.
\textsc{Anion} extends \textsc{Atomic} by negating the \emph{if} event and
manually annotating a new \emph{then} event; however, it does not consider negating the \emph{then} event.
Unlike these approaches, we incorporate negation into
the \emph{if} event, the \emph{then} event, and both.
Specifically, given a commonsense triple <\textit{A}, \textit{R}, \textit{B}> from \textsc{Atomic},
where \textit{A} is the \emph{if} event, \textit{R} is the relation, and \textit{B} is the \emph{then} event,
we leverage an LLM to add the logical negation cue \emph{not} to \textit{A}, \textit{B}, and both, generating three new triples:
<\textit{$\neg$A}, \textit{R}, \textit{B}>, <\textit{A}, \textit{R}, \textit{$\neg$B}>, and <\textit{$\neg$A}, \textit{R}, \textit{$\neg$B}>.
For a commonsense triple <\textit{$\neg$A}, \textit{R}, \textit{B$^\prime$}> from \textsc{Anion},
where \textit{B$^\prime$} is a new \emph{then} event distinct from \textit{B},
we negate only \textit{B$^\prime$} to generate
<\textit{$\neg$A}, \textit{R}, \textit{$\neg$B$^\prime$}>.
We do not negate \textit{$\neg$A}, as doing so would yield the same \emph{if} event as in \textsc{Atomic}.
Note that we use only the training split to generate these triples.

To perform negation, we add the negation cue \emph{not} to (1) the main verb of the event
(e.g., ``the person \textbf{does not} take a picture'', ``\textbf{not} look at the picture''),
or (2) the modifier when the verb is absent
(e.g., ``\textbf{not} excited'').
Using manually curated exemplars, we prompt Llama 3.1 70B to generate the negated events. 
A manual evaluation of 200 instances confirms 99\% syntactic correctness.

\subsection{Validating New Triples}
\label{ss:validating_new_triples}
It remains unknown whether the automatically generated triples with negation align with or conflict with commonsense knowledge.
We use \textit{Valid} and \textit{Invalid} to denote these cases, respectively.
In addition, we observe two scenarios when the triples are considered \textit{Ambiguous}: 
(1) the validity is context-dependent---for example, \textit{``If the sun is not shining, then it is daytime''} can
be either \textit{Valid} or \textit{Invalid} because \textit{``the sun is not shining''} could indicate heavy
clouds during daytime or simply nighttime; 
(2) the triple lacks a clear causal connection and thus has ambiguous semantics---for example, \textit{``If Person A does not get a gift, then it causes Person A to feel joy.''}

\begin{table*}
  \centering
  \small
  \begin{tabular}{l r@{\,}r r@{\,}r r@{\,}r r@{\,}r r@{\,}r}
\toprule
\multirow{3}{*}{Dataset}  & \multicolumn{2}{c}{\multirow{3}{*}{\# Triples (\%)}} & \multicolumn{2}{c}{\# w/o Neg. (\%)} & \multicolumn{6}{c}{\# with Negation (\%)} \\
\cmidrule(lr){4-5} \cmidrule(lr){6-11}
                          &        &         & \multicolumn{2}{c}{$\langle A, R, B \rangle$} & \multicolumn{2}{c}{$\langle \neg A, R, B \rangle$} & \multicolumn{2}{c}{$\langle A, R, \neg B \rangle$} & \multicolumn{2}{c}{$\langle \neg A, R, \neg B \rangle$} \\
\midrule
Existing Corpora                       \\
~~~\textsc{Atomic}$^\ast$          &   449k &         &  449k &         & \multicolumn{2}{c}{---} & \multicolumn{2}{c}{---} & \multicolumn{2}{c}{---} \\
~~~\textsc{Anion}$^\ast$           &   142k &         & \multicolumn{2}{c}{---} &  142k &         & \multicolumn{2}{c}{---} & \multicolumn{2}{c}{---} \\
\addlinespace
Our Corpora         &        &         &       &         &       &         &       &         &      &         \\
~~~$\neg$\textsc{Atomic}  & 1,798k &(100.0) &  449k &(100.0) &  449k &(100.0) &  449k &(100.0) & 449k &(100.0) \\
~~~~~~~\emph{Valid}       &   681k & (37.9) &  376k & (83.7) &   47k & (10.5) &   42k &  (9.2) & 216k & (48.0) \\
~~~~~~~\emph{Invalid}     &   463k & (25.8) &    8k &  (2.0) &  128k & (28.5) &  286k & (63.6) &  41k &  (9.1) \\
~~~~~~~\emph{Ambiguous}   &   652k & (36.3) &   64k & (14.3) &  274k & (61.0) &  122k & (27.2) & 192k & (42.9) \\
\addlinespace
~~~$\neg$\textsc{Anion}   &   285k &(100.0) & \multicolumn{2}{c}{---} &  142k &(100.0) & \multicolumn{2}{c}{---} & 142k &(100.0) \\
~~~~~~~\emph{Valid}       &   104k & (36.4) & \multicolumn{2}{c}{---} &   77k & (53.9) & \multicolumn{2}{c}{---} &  27k & (18.9) \\
~~~~~~~\emph{Invalid}     &    46k & (16.1) & \multicolumn{2}{c}{---} &    8k &  (5.6) & \multicolumn{2}{c}{---} &  38k & (26.6) \\
~~~~~~~\emph{Ambiguous}   &   135k & (47.5) & \multicolumn{2}{c}{---} &   58k & (40.5) & \multicolumn{2}{c}{---} &  77k & (54.5) \\
\addlinespace
Benchmark ($\neg$\textsc{Atomic})&  7,200 &(100.0) & 1,800 &(100.0) & 1,800 &(100.0) & 1,800 &(100.0) &1,800 &(100.0) \\
~~~~~~~\emph{Valid}       &  2,329 & (32.3) & 1,287 & (71.5) &   113 &  (6.3) &   118 &  (6.5) &  811 & (45.1) \\
~~~~~~~\emph{Invalid}     &  2,150 & (29.9) &    41 &  (2.3) &   823 & (45.7) & 1,184 & (65.8) &  102 &  (5.6) \\
~~~~~~~\emph{Ambiguous}   &  2,721 & (37.8) &   472 & (26.2) &   864 & (48.0) &   498 & (27.7) &  887 & (49.3) \\
\bottomrule
\end{tabular}

  \caption{
    Statistics of commonsense triples in existing corpora, our corpora, and the benchmark.
    \textsc{Atomic} contains no triples with negation, 
    and \textsc{Anion} negates only the \emph{if} event.
    $^\ast$ denotes subsets of the datasets: training splits from each 
  corpus, excluding underspecified triples from \textsc{Atomic} and using only the logical negation split from 
  \textsc{Anion}.}
  \label{t:analysis_example}
\end{table*}

Recent work leverages LLM-as-a-judge to evaluate LLM outputs
such as synthetic datasets~\cite{li2024generation}.
However, we demonstrate that state-of-the-art LLMs, including GPT-4o and Claude Sonnet 4,
lack the ability to evaluate the validity of commonsense knowledge triples with negation.
We address this limitation by training a task-specific LLM to automatically validate commonsense triples.

We train LLMs using supervised fine-tuning with three types of data:
(1) \textit{Valid} triples from existing commonsense corpora (\textsc{Atomic} and \textsc{Anion}),
(2) \textit{Ambiguous} triples created by randomly combining \emph{if} events, \emph{then} events, and relations from existing triples, similar to~\cite{fang-etal-2022-pseudoreasoner},
and (3) \textit{Invalid} triples constructed by prompting GPT-4o to generate \emph{then} events from existing \emph{if} events.
Note that the training data are considered noisy without any manual inspection.
For evaluation, we construct a benchmark comprising 200 triples per relation type from \textsc{Atomic} 
along with their three negated variations (7,200 triples total).
We recruit two annotators to label each triple as \textit{Valid}, \textit{Invalid}, or \textit{Ambiguous}, 
achieving an inter-annotator agreement of 0.62, indicating substantial agreement~\cite{10.1162/coli.07-034-R2}.
Note that the training and evaluation data do not overlap,
as they are sampled from the original training and test splits, respectively.
Appendix~\ref{app:annotation_details} provides more details on benchmark creation and annotation.

We experiment with Llama 3.1 8B and 70B using three training data variations---differing only in 
whether \textit{Valid} and \textit{Ambiguous} triples come from \textsc{Atomic}, \textsc{Anion}, or both;
\emph{Invalid} triples are always synthesized via an LLM.
Regardless of the training corpus source,
the models are trained on 5,400 training triples (200 triples per relation per label)
using QLoRA~\cite{dettmers2023qlora} with 4-bit quantization.
Both training and evaluation data are verbalized from commonsense triples to natural language if-then statements.
Further experimental details including the verbalization mapping can be found in Appendix~\ref{app:validating_triples}.

Table~\ref{t:validation_results} reports the validation results on our benchmark.
Claude Sonnet 4 outperforms the other LLMs, though it achieves only 0.56 F1 score overall.
Fine-tuned Llama 3.1 70B model outperforms all proprietary models
(F1 score: 0.63 vs. 0.56; Accuracy: 0.64 vs. 0.56).
Llama 3.1 8B shows lower performance, as expected for a smaller model.
We consider precision for \textit{Valid} and \textit{Invalid} triples the more critical metric, as only triples with these two labels are used for training (Section~\ref{s:reasoning_with_negation}).
Lower recall is tolerable since it only reduces training set size.
Our best LLM judge achieves 0.70 and 0.79 precision for \textit{Valid} and \textit{Invalid} triples, respectively.
Moreover, empirical experiments (Section~\ref{s:experiment}) demonstrate that
the commonsense corpora synthesized using our LLM judge effectively improve LLMs' negation understanding.

\subsection{Analysis of Commonsense Knowledge Corpora with Negation}

We develop two new corpora with negation, $\neg$\textsc{Atomic} and $\neg$\textsc{Anion}, 
which are validated by our LLM judge (Table~\ref{t:validation_results}, the last row) with three labels,
\emph{Valid}, \emph{Invalid}, and \emph{Ambiguous}.
Table \ref{t:analysis_example} presents statistics for the existing corpora (\textsc{Atomic} and \textsc{Anion}), 
our corpora, and our benchmark.

We use only a subset of the existing corpora:
the training split, excluding underspecified triples (e.g., PersonX sees \_\_\_ in the water) from 
\textsc{Atomic} and using only the logical negation split from \textsc{Anion}. 
Our approach augmenting commonsense triples with negation is effective:
37.9\% of the commonsense triples from $\neg$\textsc{Atomic} (36.4\% from $\neg$\textsc{Anion}) are identified as \emph{Valid}, aligning with commonsense knowledge,
while fewer triples are identified as \emph{Invalid} ($\neg$\textsc{Atomic}: 25.8\%; $\neg$\textsc{Anion}: 16.1\%).
More importantly, they are augmented with negation.
As we demonstrate in Section~\ref{s:experiment}, both \emph{Valid} and \emph{Invalid} triples improve LLMs' ability to understand negation.

\section{Commonsense Knowledge with Negation for Downstream Tasks}
\label{s:reasoning_with_negation}
We have presented a novel method to automatically augment
existing commonsense knowledge corpora with negation. 
Our two corpora, $\neg$\textsc{Atomic} and $\neg$\textsc{Anion},
contain over 2M commonsense knowledge triples augmented with negation, validated by an LLM judge.
Beyond this contribution, we demonstrate that our corpora improve LLMs' ability to understand negation. 
Specifically, we evaluate on five benchmarks across three tasks:
question answering (QA), natural language inference (NLI),
and information retrieval (IR).

\subsection{Benchmarks Evaluating LLMs' Negation Understanding}

\textbf{CondaQA} \cite{ravichander-etal-2022-condaqa} is a contrastive QA dataset requiring understanding of negation cues 
in passages to answer questions.
Each question is paired with a passage containing the answer,
with answers being either \emph{Yes}, \emph{No}, \emph{Don't know}, a span in the question, or a span in the context.
Following~\citet{ravichander-etal-2022-condaqa},
we evaluate CondaQA using two metrics: \emph{accuracy} and \emph{group consistency}.
The term \emph{group} refers to the original passage and either all three or one of the edited passages.
\emph{Group consistency} measures the percentage of questions answered correctly for all passages in a group,
which is arguably more important than accuracy, as robustness against negation requires correctly
answering questions with both original and negated passages.

\textbf{NLI with Negation} is introduced by
\citet{hossain-etal-2020-analysis}. It contains three NLI benchmarks
developed from existing benchmarks: RTE \cite{10.1007/11736790_9}, SNLI \cite{bowman-etal-2015-large}, and MNLI \cite{williams-etal-2018-broad}.
The authors show that existing NLI benchmarks contain few negation cues
and develop new benchmarks
by negating the main verbs in premises and hypotheses to create three new pairs from each original pair.
The new pairs are manually annotated to obtain labels.

\textbf{NevIR}~\cite{weller-etal-2024-nevir} addresses the weakness of neural information retrieval (IR) models in understanding negation.
The dataset is constructed using contrastive query-document pairs from CondaQA,
where each pair of queries and documents is nearly identical except for a crucial negation.
An IR model is expected to rank documents based on queries by correctly understanding negation.
Pairwise accuracy serves as the evaluation metric: the model must correctly rank documents for both queries, 
flipping the ranking when given the negated query.
Although the dataset's primary purpose is to evaluate IR model performance,
it is essentially a binary classification task, as only two documents are provided for each query.
Still, it is considered a challenging benchmark as models are required to understand negation from long context (documents).

\subsection{Training LLMs with Commonsense Knowledge with Negation}
\label{ss:creating}
Although over half of our generated triples are identified as \textit{Invalid} (Table~\ref{t:analysis_example}),
meaning they conflict with commonsense knowledge,
they remain useful for teaching LLMs to understand negation.
Specifically, we design a training objective that enables models to learn from both \textit{Valid} and \textit{Invalid} triples,
and how negating the \emph{if} or \emph{then} event affects triple validity.
We construct training corpora by selecting contrastive triples from $\neg$\textsc{Atomic} following the patterns below.
We select triples if:
\begin{compactitem}
    \item the original triple <\textit{A}, \textit{R}, \textit{B}> is \emph{Valid},
    <\textit{$\neg$A}, \textit{R}, \textit{B}> is \emph{Invalid},
    and <\textit{A}, \textit{R}, \textit{$\neg$B}> is \emph{Valid};
    \item the original triple <\textit{A}, \textit{R}, \textit{B}> is \emph{Valid},
    <\textit{$\neg$A}, \textit{R}, \textit{B}> is \emph{Valid},
    and <\textit{A}, \textit{R}, \textit{$\neg$B}> is \emph{Invalid};
    \item the original triple <\textit{A}, \textit{R}, \textit{B}> is \emph{Invalid},
    <\textit{$\neg$A}, \textit{R}, \textit{B}> is \emph{Valid},
    and <\textit{A}, \textit{R}, \textit{$\neg$B}> is \emph{Invalid}; or
    \item the original triple <\textit{A}, \textit{R}, \textit{B}> is \emph{Invalid},
    <\textit{$\neg$A}, \textit{R}, \textit{B}> is \emph{Invalid},
    and <\textit{A}, \textit{R}, \textit{$\neg$B}> is \emph{Valid}.
\end{compactitem}

For $\neg$\textsc{Anion}, we select triples where the original and negated triples have different labels.
\begin{table*}[t!]
  \centering
  \small
  \begin{tabular}{l r r@{ }l rrrr}
\toprule
                                                              &            &          &                  & \multicolumn{4}{c}{Group Consistency} \\
\cmidrule(lr){5-8}
                                                              & \# Params. & Accuracy & ($\Delta$\%)     &  All &  Par. &  Sco. &  Aff. \\
\midrule
\textbf{Fully supervised}                                     \\
~~~UnifiedQA-v2-large~\cite{ravichander-etal-2022-condaqa}    & 770M       &     66.7 &&                   30.2 &  64.0 &  43.7 &  46.5 \\
~~~RoBERTa-large                                              & 355M       &     64.9 &&                   29.6 &  61.9 &  41.4 &  45.8 \\
~~~~~~Pre-trained with \\
~~~~~~~~~\textsc{Atomic} + \textsc{Anion}                                    &            &       67.0 & (+3.2)          & 32.9 &  65.6 &  46.3 &  48.1 \\
~~~~~~~~~Best of ($\neg$\textsc{Atomic}, $\neg$\textsc{Anion})                        &&        \textbf{68.5} & (+5.5)   & \textbf{34.3} & \textbf{66.4} & \textbf{47.6} & \textbf{50.1} \\
\midrule
\textbf{In-context learning}                                            \\
~~~\textbf{Zero-shot}                                            \\
~~~~~~OpenAI o1                                                  & ---        &     65.3 &&                   24.9 &  67.4 &  43.8 & 38.6 \\
~~~\textbf{Few-shot}                                             \\
~~~~~~InstructGPT + COT~\cite{ravichander-etal-2022-condaqa}     & ---        &     66.3 &&                   27.3 &  64.2 &  45.1 &  44.9 \\
~~~~~~GPT-4o                                                     & ---        &     72.9 &&                   34.4 &  78.7 &  52.6 &  47.9 \\
~~~~~~Llama 3.1 70B                                              & 70B        &     77.5 &&                   43.3 &  83.7 &  61.8 &  54.9 \\
~~~~~~Llama 3.1 8B                                               & 8B         &     68.7 &&                   31.5 &  69.0 &  48.1 &  44.4 \\
~~~~~~~~~Pre-trained with                                \\
~~~~~~~~~~~~\textsc{Atomic} + \textsc{Anion}                                      
                                                              &&               67.6 &(-1.6) & 27.5 & 69.5 & 44.3 & 41.4 \\
~~~~~~~~~~~~Best of ($\neg$\textsc{Atomic},$\neg$\textsc{Anion})                           
                                                              &&       \textbf{71.5}  & (+4.1)$^\ast$   & \textbf{33.5} & \textbf{72.9} & \textbf{48.7} & \textbf{46.4} \\
   
\addlinespace

~~~~~~Qwen2 7B                                                   & 7B         &     65.1 &&                   24.9 &  65.9 &  39.7 &  39.2 \\
~~~~~~~~~Pre-trained with                                \\
~~~~~~~~~~~~\textsc{Atomic} + \textsc{Anion}                     &&                 64.1 & (-1.5) & 22.8 & 65.2 & 38.4 & 37.7 \\
~~~~~~~~~~~~Best of ($\neg$\textsc{Atomic}, $\neg$\textsc{Anion})                     
                                                              &&       69.7  & (+7.1)$^\ast$   & 32.7 & 71.2 & 46.3 & 45.5 \\
\bottomrule
\end{tabular}
  \caption{
      Results evaluating CondaQA 
      with two settings: fully supervised (top) and in-context learning (bottom). The best result for each model is in bold.
      Delta ($\Delta$) indicates the percent change in accuracy compared to the off-the-shelf model.
      An asterisk $^\ast$ indicates a statistically significant
      improvement (McNemar’s test~\cite{mcnemar1947note}, $p < 0.05$) over both the off-the-shelf model and the model trained with existing corpora.
    }
  \label{t:condaqa_evaluation}

\end{table*}
This approach balances the distribution of \emph{Valid} and \emph{Invalid} triples
and, more importantly,
highlights two patterns for models to compare:
(1) negating either the \emph{if} or \emph{then} event keeps the new triple's validity (either \emph{Valid} or \emph{Invalid}), and
(2) negating either the \emph{if} or \emph{then} event flips the new triple's validity.
We do not include <\textit{$\neg$A}, \textit{R}, \textit{$\neg$B}> as double negations result in more complex semantics.
The final training set consists of 89k triples from $\neg$\textsc{Atomic} and 76k triples from $\neg$\textsc{Anion}.
Finally, the model is trained to predict whether a triple is \emph{Valid} or \emph{Invalid}, using the validation results (Section~\ref{ss:validating_new_triples})
as ground truth.

As a baseline, we also construct a training dataset without augmented negation.
This is done by sampling the commonsense triples
from \textsc{Atomic} and \textsc{Anion} as \emph{Valid} instances, 
while reusing the \emph{Invalid} triples synthesized by an LLM (Section~\ref{ss:validating_new_triples}).
Although \textsc{Anion} contains triples with negated \emph{if} events,
they differ significantly from our corpora as they do not preserve the original \emph{then} event.
Note that we always sample the same number of triples from \textsc{Atomic} and \textsc{Anion} as $\neg$\textsc{Atomic} and $\neg$\textsc{Anion}.

\section{Experiments}
\label{s:experiment}
We evaluate our corpora on negation understanding using an encoder-based model (RoBERTa-large) and two LLMs (Llama 3.1 8B and Qwen2 7B) across three tasks:
(1) CondaQA, a QA task (Section~\ref{ss:condaqa});
(2) NLI with Negation, an NLI task (Section~\ref{ss:nli});
and (3) NevIR, an IR task (Section~\ref{ss:nevir}).

All three models are first pre-trained on our commonsense corpora, 
$\neg$\textsc{Atomic}, $\neg$\textsc{Anion}, or both. 
As a baseline, models are pre-trained on existing corpora (\textsc{Atomic} and \textsc{Anion}).
We then evaluate on downstream tasks in two settings:
(1) fully-supervised fine-tuning on the downstream task's training split for encoder-based models,
and (2) zero-shot and few-shot in-context learning for LLMs.
Following standard practices, we use a zero-shot prompt with the OpenAI o1 model and a few-shot prompt with GPT-4o and open-source LLMs.
Pre-trained LLMs are trained and evaluated locally using QLoRA~\cite{dettmers2023qlora} due to computational resource limitations.
Appendix~\ref{app:evaluation_details} reports experimental details including the prompts and hyperparameters.

\begin{table*}[t!]
  \centering
  \small
  \begin{tabular}{l r r@{}l r@{}l r@{}l r@{}l}
\toprule
                                                       & \multirow{3}{*}{\# Params.} & \multicolumn{6}{c}{NLI with Negation}       & \multirow{3}{*}{NevIR} \\
\cmidrule(lr){3-8}
                                                       &      &          RTE-Neg &&         SNLI-Neg &&         MNLI-Neg          \\
\midrule
\textbf{Fully supervised}                              \\
~~~BERTNOT~\cite{hosseini-etal-2021-understanding}     & 110M &             74.5 &&             46.0 &&             60.9 &&   --- \\
~~~RoBERTa-large-NSP~\cite{rezaei-blanco-2025-making}  & 355M &             87.2 &&             56.5 &&             69.9 &&   --- \\
~~~stsb-roberta-large~\cite{weller-etal-2024-nevir}    & 355M &              --- &&              --- &&              --- &&  24.9 \\
~~~MonoT5 3B~\cite{nogueira-etal-2020-document}        & 3B   &              --- &&              --- &&              --- &&  50.6 \\
~~~RoBERTa-large                                       & 355M &             84.7 &&             56.0 &&             \textbf{69.9} &&  24.5 \\
~~~~~~Pre-trained with           \\             
~~~~~~~~~\textsc{Atomic} + \textsc{Anion}              &&                   86.2 &&             56.5 &&             69.4 &&  29.1 \\
~~~~~~~~~Best of ($\neg$\textsc{Atomic}, $\neg$\textsc{Anion})  && \textbf{88.1} &$^\ast$&   \textbf{58.3}  &&      69.7 && \textbf{34.3} &$^\ast$ \\
\midrule

\textbf{In-context learning}                                     \\
~~~\textbf{Zero-shot}                                     \\
~~~~~~OpenAI o1                                           & ---  &             87.5 &&             75.9 &&             74.6 &&  59.7 \\
~~~\textbf{Few-shot}                                      \\
~~~~~~GPT-4o                                              & ---  &             86.9 &&             74.8 &&             75.0 &&  61.7 \\
~~~~~~Llama 3.1 70B                                       & 70B  &             78.9 &&             69.1 &&             65.9 &&  58.8 \\
~~~~~~Llama 3.1 8B                                        & 8B   &             60.0 &&             54.4 &&             47.0 &&  30.6 \\
~~~~~~~~~Pre-trained with                         \\
~~~~~~~~~~~~\textsc{Atomic} + \textsc{Anion}                    &&                 65.3 &&             57.5 &&             51.1 &&  37.9 \\
~~~~~~~~~~~~Best of ($\neg$\textsc{Atomic}, $\neg$\textsc{Anion})&&                81.3 &$^\ast$&      68.2  &$^\ast$&     63.9&$^\ast$& \textbf{42.2}&$^\ast$ \\
\addlinespace
~~~~~~Qwen2 7B                                            & 7B   &             71.7 &&             60.7 &&             59.5 &&  29.8 \\
~~~~~~~~~Pre-trained with                         \\
~~~~~~~~~~~~\textsc{Atomic} + \textsc{Anion}                                     &&                 78.3 &&             66.5 &&             63.0 &&  33.8 \\
~~~~~~~~~~~~Best of ($\neg$\textsc{Atomic}, $\neg$\textsc{Anion})       &&                 \textbf{82.3} &$^\ast$&     \textbf{72.4} &$^\ast$&  \textbf{67.5} &$^\ast$&  39.8&$^\ast$ \\

\bottomrule
\end{tabular}

  \caption{
      Results evaluating three NLI benchmarks with negation cues (accuracy) and NevIR (pairwise accuracy) in two settings: 
      (1) fully supervised (top) and (2) in-context learning (bottom). The best results for each model are in bold.
      An asterisk $^\ast$ indicates a statistically significant improvement (McNemar’s test~\cite{mcnemar1947note}, $p < 0.05$) over both the off-the-shelf model and the model trained with existing corpora.
    }
  \label{t:nli_nevIR_evaluation}
\end{table*}

\subsection{Evaluating with CondaQA}
\label{ss:condaqa}
Table~\ref{t:condaqa_evaluation} reports results on CondaQA using two metrics: \emph{accuracy} and \emph{group consistency}~\cite{ravichander-etal-2022-condaqa}.
For pre-trained models, we report only the best results among three corpus configurations. Appendix~\ref{app:full_results} provides the complete results. 

RoBERTa-large benefits from pre-training on both existing and our commonsense corpora, with our corpora yielding higher performance.
Pre-training on our corpora also outperforms UnifiedQA-v2-large~\cite{ravichander-etal-2022-condaqa}, despite the latter being a larger model (770M vs. 355M).
Surprisingly, proprietary models (OpenAI o1 and GPT-4o) yield worse results than the less powerful open-source Llama 3.1 70B (65.3 vs. 72.9 vs. 77.5).
We hypothesize that OpenAI o1 with a few-shot prompt may achieve higher performance, though at significantly greater token cost.
The two LLMs pre-trained on our corpora consistently outperform the base model, with Llama 3.1 8B outperforming Qwen2 7B.
Pre-trained Llama 3.1 8B even achieves competitive performance with GPT-4o (71.5 vs. 72.9).
More importantly, most improvements from our corpora are statistically significant over both the off-the-shelf baseline and models trained with existing corpora
(indicated with $^\ast$ in Table~\ref{t:condaqa_evaluation}).
In contrast, pre-training on existing corpora yields worse results than the off-the-shelf baseline.
Note that the baseline with existing corpora uses a single configuration: \textit{Valid} triples from \textsc{Atomic} and \textsc{Anion}, and synthesized \textit{Invalid} triples.

Importantly, pre-training on our corpora does not degrade performance with non-negation instances.
The Affirmative Edit setting in CondaQA (column Aff.) requires the model to reason over passages with negation removed, and we observe no drop in performance, indicating that pre-trained models remain capable of handling affirmative text.
Appendix~\ref{app:commonsenseqa} provides extra results with CommonsenseQA~\cite{talmor-etal-2019-commonsenseqa}, a standard commonsense benchmark, further confirming this finding.

\begin{table*}[t!]
  \centering
  \small
  \begin{tabular}{l  r r r r r}
\toprule
                                  & \multirow{2}{*}{CondaQA}  & \multicolumn{3}{c}{NLI with Negation}       & \multirow{3}{*}{NevIR} \\
  \cmidrule(lr){3-5}
                                    & (Accuracy)  &                              RTE-Neg  &         SNLI-Neg &         MNLI-Neg                 \\
\midrule
Llama 3.1 8B                     &               68.7   &            60.0   &             54.4 &         47.0      &         30.6         \\
~~~Pre-trained with       \\
~~~~~~$\neg$\textsc{Atomic}      &               70.6   &            81.2   &    \textbf{68.5} &         63.8      &         38.1          \\
~~~~~~~~~<\textit{$\neg$A}, \textit{R}, \textit{B}>  
                                    &               69.3   &            76.8   &            64.0  &         53.5      &         34.3 \\
~~~~~~~~~<\textit{A}, \textit{R}, \textit{$\neg$B}>  
                                    &               67.6   &            74.0   &            65.5  &         54.6      &         35.6 \\

\addlinespace

~~~~~~$\neg$\textsc{Anion}       &      \textbf{71.3}   &    \textbf{81.3}   &            68.2  &  \textbf{63.9}    &  \textbf{42.2}\\
~~~~~~~~~<\textit{$\neg$A}, \textit{R}, \textit{$\neg$B}> 
                                    &               68.7   &             70.2   &            62.9  &         51.6      &         35.3 \\
\addlinespace

Qwen2 7B                         &               65.1   &             71.7 &             60.7  &             59.5 & 29.8          \\
~~~Pre-trained with               \\
~~~~~~$\neg$\textsc{Atomic}      &       \textbf{69.7}  &     \textbf{82.3}&      \textbf{72.4}&    \textbf{67.5} &  \textbf{39.8}\\

~~~~~~~~~<\textit{$\neg$A}, \textit{R}, \textit{B}>  
                                    &               65.8   &     74.1          &            65.5  &         65.1     &         35.8 \\
~~~~~~~~~<\textit{A}, \textit{R}, \textit{$\neg$B}>  
                                    &               64.3   &     75.7          &            68.3  &         62.4     &         34.4 \\
\addlinespace

~~~~~~$\neg$\textsc{Anion}       &               69.2   &     78.9          &            66.1  &         62.5     &         36.3 \\
~~~~~~~~~<\textit{$\neg$A}, \textit{R}, \textit{$\neg$B}> 
                                    &               65.4   &      71.0         &            54.3  &         52.5     &         30.1 \\

\bottomrule
\end{tabular}

  \caption{
      Ablation results evaluating CondaQA, NLI, and NevIR, with pre-trained LLMs on different types of negations: \textit{if} event, \textit{then} event, or both. 
}
  \label{t:negation_ablation_results}
\end{table*}

\subsection{Evaluating with NLI Benchmarks}
\label{ss:nli}
We further evaluate on three NLI benchmarks with negation: RTE-Neg, SNLI-Neg, and MNLI-Neg.
We include additional fully-supervised baselines: BERTNOT~\cite{hosseini-etal-2021-understanding}, 
RoBERTa-large-NSP~\cite{rezaei-blanco-2025-making}, stsb-roberta-large~\cite{weller-etal-2024-nevir}, and MonoT5 3B~\cite{nogueira-etal-2020-document}. 
Table~\ref{t:nli_nevIR_evaluation} (NLI with Negation) reports results using \emph{accuracy}.
Pre-training RoBERTa-large on our corpora yields the best results among all models and outperforms all baselines. However, only one result yields statistically significant improvement 
over the off-the-shelf baseline.
We hypothesize this is due to potential overfitting---models are further fine-tuned on each benchmark's training split when the tasks are as simple as classification.
Moreover, the off-the-shelf models already achieve strong results.

In-context learning with LLMs shows more expected trends.
The OpenAI o1 model achieves the highest results overall, followed by GPT-4o.
LLMs pre-trained on our corpora demonstrate statistically significant improvements over
both the off-the-shelf baseline and models pre-trained on existing corpora across all three benchmarks.
Notably, Qwen2 7B even outperforms the larger Llama 3.1 70B model.
Unlike CondaQA, pre-training on existing corpora also yields benefits, despite being significantly smaller than our corpora.
Note that the existing corpora include negated triples from \textsc{Anion}, which benefit models' negation understanding on the simpler NLI task. 

\begin{table*}[t!]
  \centering
  \small
  \begin{tabular}{l r r r r r r}
\toprule
                                  & \multirow{2}{*}{\# Triples}  & \multirow{2}{*}{CondaQA}  & \multicolumn{3}{c}{NLI with Negation}       & \multirow{3}{*}{NevIR} \\
  \cmidrule(lr){4-6}
                                  &               & (Accuracy)  &                              RTE-Neg  &         SNLI-Neg &         MNLI-Neg                 \\
\midrule
Llama 3.1 8B                     & n/a           &        68.7 &            60.0 &            54.4  &         47.0     &         30.6 \\
~~~Pre-trained with              \\
~~~~~~$\neg$\textsc{Atomic} + $\neg$\textsc{Anion} 
                                & 1K + 1K             &        67.7 &            70.8 &            57.1  &         53.5     &         31.8 \\
                                & 10K + 10K           &        69.3 &            81.0 &            67.7  &         62.9     &         34.5 \\
                                & 89K + 76K          &        71.5 &            81.3 &            67.9  &         63.7     &         38.2 \\

\addlinespace

Qwen2 7B                        & n/a           &        65.1 &            71.7 &            60.7  &         59.5     &         29.8 \\
~~~Pre-trained with              \\
~~~~~~$\neg$\textsc{Atomic} + $\neg$\textsc{Anion}
                                & 1K + 1K           &        65.2 &            76.0 &            65.2  &         63.1     &         34.9 \\
                                & 10K + 10K         &        66.6 &            81.7 &            72.2  &         67.4     &         36.0 \\
                                & 89K + 76K          &        66.6 &            82.1 &            72.2  &         67.5     &         38.5 \\

\bottomrule
\end{tabular}

  \caption{
      Ablation on training dataset size. We train Llama 3.1 8B and Qwen2 7B with two subsets of $\neg$\textsc{Atomic} + $\neg$\textsc{Anion}: (1) 2K triples (1K from each dataset) and (2) 20K triples (10K each).
      Training with the 2K subset yields significantly worse performance than the complete dataset, while training with the 20K subset shows substantial improvement and tends to saturate on simpler tasks such as NLI.
}
  \label{t:training_size_ablation_study}
\end{table*}

\subsection{Evaluating with NevIR}
\label{ss:nevir}

Following~\citet{weller-etal-2024-nevir}, we perform fully-supervised fine-tuning with RoBERTa-large on STS-B~\cite{cer-etal-2017-semeval} instead of NevIR's training split.
Table~\ref{t:nli_nevIR_evaluation} (NevIR) reports the results using pairwise accuracy---the model needs to correctly rank documents for both queries (with and without negation).
Although NevIR is a retrieval task, it only requires ranking between two documents for each query and essentially becomes binary classification. 
In fact, RoBERTa-large pre-trained on our corpora outperforms a customized IR model baseline (stsb-roberta-large~\cite{weller-etal-2024-nevir}, 34.3 vs. 24.9).
Again, pre-training on our corpora yields statistically significant improvements over
off-the-shelf models and models trained with existing corpora.
The results are consistent across all three models.
All our models significantly underperform OpenAI o1, GPT-4o, and even MonoT5 3B~\cite{nogueira-etal-2020-document}.
We hypothesize that pre-training a retrieval model with our corpora as a starting point would yield greater benefits.

\subsection{Ablation Studies}
\label{ss:ablations}

\paragraph{Pre-training with Individual Negation Types}
We further investigate whether training with individual negation types contributes differently to performance improvements.
For $\neg$\textsc{Atomic}, we compare the complete corpus against training with triples that only negate the \emph{if} event
(<\textit{$\neg$A}, \textit{R}, \textit{B}>) 
or the \emph{then} event (<\textit{A}, \textit{R}, \textit{$\neg$B}>).
For $\neg$\textsc{Anion}, we compare against training with triples that negate both events (<\textit{$\neg$A}, \textit{R}, \textit{$\neg$B}>).

Table~\ref{t:negation_ablation_results} reports the results with two LLMs. 
Note that results with the complete corpus are not directly comparable to Table~\ref{t:condaqa_evaluation} and~\ref{t:nli_nevIR_evaluation}, which report only the best configuration.
Training with individual negation types consistently underperforms the complete corpus across all benchmarks, 
demonstrating that the patterns within our corpora are critical (Section~\ref{ss:creating}).
Notably, negating the \emph{if} event alone yields higher results than the off-the-shelf model,
suggesting that certain negation types still provide partial benefits for negation understanding.

For $\neg$\textsc{Anion}, training with triples negating both events yields substantially worse performance than the complete corpus.
We hypothesize that double negation alone is insufficient, as models benefit from first learning simpler single-negation patterns before generalizing to more complex negations.

\paragraph{Pre-training with Different Data Sizes}

Our full training dataset consists of 89k triples from $\neg$\textsc{Atomic} and 76k triples from $\neg$\textsc{Anion} (Section~\ref{ss:creating}).
We conduct two ablations on the effect of training data size on downstream task performance: (1) training with a 2,000-triple subset (randomly sampled and balanced with 1,000 triples each from $\neg$\textsc{Atomic} and $\neg$\textsc{Anion}),
and (2) training with a 20,000-triple subset (10,000 each).
Table~\ref{t:training_size_ablation_study} reports the results compared to off-the-shelf models and models trained with the complete dataset.
Similarly, Table~\ref{t:condaqa_evaluation} and Table~\ref{t:nli_nevIR_evaluation} only report the best result among $\neg$\textsc{Atomic}, $\neg$\textsc{Anion}, or both.
Appendix~\ref{app:additional_results} reports comparable results with [$\neg$\textsc{Atomic} + $\neg$\textsc{Anion}].

Training with only 2K triples already improves over the off-the-shelf models on NLI tasks, though the gains are modest.
Scaling to 20K triples yields substantial improvements, with NLI performance approaching that of the full dataset (e.g., RTE improves from 70.8 to 81.0 for Llama 3.1 8B, compared to 81.3 with the full training set).
However, NevIR continues to benefit from additional training data beyond 20K (34.5 $\rightarrow$ 38.2 for Llama 3.1 8B, 36.0 $\rightarrow$ 38.5 for Qwen2 7B),
suggesting that more complex negation reasoning tasks benefit from larger training sets.

\paragraph{Pre-training with Randomly Labelled Data}
We further validate the importance of our LLM judge by training models on randomly labelled data.
Training with randomly labelled data substantially degrades performance compared to LLM-validated data, particularly on NevIR where performance drops well below the off-the-shelf baseline.
Due to space limitations, we report the full results and analyses in Appendix~\ref{app:random_label}; 
Appendix~\ref{app:error_analysis} further provides an error analysis categorizing improvements by negation type~\cite{hossain-etal-2020-analysis} 
and case studies illustrating specific reasoning patterns improved by our approach.

\section{Conclusion}
Negation and commonsense knowledge are both common and important in human language.
Previous work shows that models struggle when negation appears in natural language understanding tasks.
However, few works have investigated commonsense knowledge with negation.

We present an approach to automatically augment existing commonsense knowledge corpora with negation,
contributing over 2M commonsense knowledge triples with negation.
We show that pre-training models with our corpora is beneficial in understanding negation.
This holds true across three models and five benchmarks. 
We further conduct ablation studies and analyses
that provide additional evidence and insights into the performance improvements.

\section*{Limitations}
We work on two existing commonsense knowledge corpora with limited focus---they
only contain \emph{if-then} relations.
It would be more comprehensive to investigate commonsense knowledge in different forms. 
In addition, we only consider the logical negation cue \emph{not}. Future work should consider various negation cues, including semantic negation.

We only experiment with two relatively small LLMs (a 7B and an 8B model) to study the benefit of our corpora for improving negation understanding. 
Due to limited computational resources, we choose not to
conduct experiments with larger LLMs (e.g., 70B).
Moreover, the LLMs are trained and evaluated with 4-bit quantization due to the same resource limitation.

NevIR was developed to evaluate information retrieval tasks involving negation.
We experiment with NevIR as a classification task using either classifier models or general-purpose LLMs.
Future work should consider models specifically designed for information retrieval.

\section*{Ethics Statement}

\textbf{Data Sources and Collection}
We collect the datasets (\textsc{Atomic}, \textsc{Anion}, CondaQA, NLI with Negation, and NevIR) via links provided by the authors.

Data from \textsc{Atomic}~\cite{sap2019atomic} is used under the Creative Commons Attribution 4.0 International License;
CondaQA~\cite{ravichander-etal-2022-condaqa} under Apache-2.0 License;
NLI with Negation~\cite{hossain-etal-2020-analysis} and
NevIR~\cite{weller-etal-2024-nevir} under MIT License.
\textsc{Anion}~\cite{jiang-etal-2021-im} does not specify its license.

\section*{Acknowledgments}
We thank the reviewers for their insightful comments.

The OpenAI Researcher Access Program provided credits
to conduct this research.

This material is based upon work supported
by the National Science Foundation under Grant
No. 2310334. Any opinions, findings, and conclusions or recommendations expressed in this material are those of the authors and do not necessarily
reflect the views of the NSF.

\bibliography{custom}

@inproceedings{fang-etal-2022-pseudoreasoner,
    title = "{P}seudo{R}easoner: Leveraging Pseudo Labels for Commonsense Knowledge Base Population",
    author = "Fang, Tianqing  and
      Do, Quyet V.  and
      Zhang, Hongming  and
      Song, Yangqiu  and
      Wong, Ginny Y.  and
      See, Simon",
    editor = "Goldberg, Yoav  and
      Kozareva, Zornitsa  and
      Zhang, Yue",
    booktitle = "Findings of the Association for Computational Linguistics: EMNLP 2022",
    month = dec,
    year = "2022",
    address = "Abu Dhabi, United Arab Emirates",
    publisher = "Association for Computational Linguistics",
    url = "https://aclanthology.org/2022.findings-emnlp.246/",
    doi = "10.18653/v1/2022.findings-emnlp.246",
    pages = "3379--3394"
}

@inproceedings{10.1145/3511808.3557484,
author = {Arnaout, Hiba and Razniewski, Simon and Weikum, Gerhard and Pan, Jeff Z.},
title = {UnCommonSense: Informative Negative Knowledge about Everyday Concepts},
year = {2022},
isbn = {9781450392365},
publisher = {Association for Computing Machinery},
address = {New York, NY, USA},
url = {https://doi.org/10.1145/3511808.3557484},
doi = {10.1145/3511808.3557484},
booktitle = {Proceedings of the 31st ACM International Conference on Information \& Knowledge Management},
pages = {37–46},
numpages = {10},
location = {Atlanta, GA, USA},
series = {CIKM '22}
}

@inproceedings{sap2019atomic,
  title={Atomic: An atlas of machine commonsense for if-then reasoning},
  author={Sap, Maarten and Le Bras, Ronan and Allaway, Emily and Bhagavatula, Chandra and Lourie, Nicholas and Rashkin, Hannah and Roof, Brendan and Smith, Noah A and Choi, Yejin},
  booktitle={Proceedings of the AAAI conference on artificial intelligence},
  volume={33},
  number={01},
  pages={3027--3035},
  year={2019}
}

@inproceedings{speer2017conceptnet,
  title={Conceptnet 5.5: An open multilingual graph of general knowledge},
  author={Speer, Robyn and Chin, Joshua and Havasi, Catherine},
  booktitle={Proceedings of the AAAI conference on artificial intelligence},
  volume={31},
  number={1},
  year={2017}
}

@article{touvron2023llama,
  title={Llama: Open and efficient foundation language models},
  author={Touvron, Hugo and Lavril, Thibaut and Izacard, Gautier and Martinet, Xavier and Lachaux, Marie-Anne and Lacroix, Timoth{\'e}e and Rozi{\`e}re, Baptiste and Goyal, Naman and Hambro, Eric and Azhar, Faisal and others},
  journal={arXiv preprint arXiv:2302.13971},
  year={2023}
}

@inproceedings{rezaei-blanco-2024-paraphrasing,
    title = "Paraphrasing in Affirmative Terms Improves Negation Understanding",
    author = "Rezaei, MohammadHossein  and
      Blanco, Eduardo",
    editor = "Ku, Lun-Wei  and
      Martins, Andre  and
      Srikumar, Vivek",
    booktitle = "Proceedings of the 62nd Annual Meeting of the Association for Computational Linguistics (Volume 2: Short Papers)",
    month = aug,
    year = "2024",
    address = "Bangkok, Thailand",
    publisher = "Association for Computational Linguistics",
    url = "https://aclanthology.org/2024.acl-short.55/",
    doi = "10.18653/v1/2024.acl-short.55",
    pages = "602--615"
}

@inproceedings{singh-etal-2023-nlms,
    title = "{NLM}s: Augmenting Negation in Language Models",
    author = "Singh, Rituraj  and
      Kumar, Rahul  and
      Sridhar, Vivek",
    editor = "Bouamor, Houda  and
      Pino, Juan  and
      Bali, Kalika",
    booktitle = "Findings of the Association for Computational Linguistics: EMNLP 2023",
    month = dec,
    year = "2023",
    address = "Singapore",
    publisher = "Association for Computational Linguistics",
    url = "https://aclanthology.org/2023.findings-emnlp.873/",
    doi = "10.18653/v1/2023.findings-emnlp.873",
    pages = "13104--13116"
}

@article{achiam2023gpt,
  title={Gpt-4 technical report},
  author={Achiam, Josh and Adler, Steven and Agarwal, Sandhini and Ahmad, Lama and Akkaya, Ilge and Aleman, Florencia Leoni and Almeida, Diogo and Altenschmidt, Janko and Altman, Sam and Anadkat, Shyamal and others},
  journal={arXiv preprint arXiv:2303.08774},
  year={2023}
}

@misc{yang2024qwen2technicalreport,
      title={Qwen2 Technical Report}, 
      author={An Yang and Baosong Yang and Binyuan Hui and Bo Zheng and Bowen Yu and Chang Zhou and Chengpeng Li and Chengyuan Li and Dayiheng Liu and Fei Huang and Guanting Dong and Haoran Wei and Huan Lin and Jialong Tang and Jialin Wang and Jian Yang and Jianhong Tu and Jianwei Zhang and Jianxin Ma and Jianxin Yang and Jin Xu and Jingren Zhou and Jinze Bai and Jinzheng He and Junyang Lin and Kai Dang and Keming Lu and Keqin Chen and Kexin Yang and Mei Li and Mingfeng Xue and Na Ni and Pei Zhang and Peng Wang and Ru Peng and Rui Men and Ruize Gao and Runji Lin and Shijie Wang and Shuai Bai and Sinan Tan and Tianhang Zhu and Tianhao Li and Tianyu Liu and Wenbin Ge and Xiaodong Deng and Xiaohuan Zhou and Xingzhang Ren and Xinyu Zhang and Xipin Wei and Xuancheng Ren and Xuejing Liu and Yang Fan and Yang Yao and Yichang Zhang and Yu Wan and Yunfei Chu and Yuqiong Liu and Zeyu Cui and Zhenru Zhang and Zhifang Guo and Zhihao Fan},
      year={2024},
      eprint={2407.10671},
      archivePrefix={arXiv},
      primaryClass={cs.CL},
      url={https://arxiv.org/abs/2407.10671}, 
}

@article{grattafiori2024llama,
  title={The llama 3 herd of models},
  author={Grattafiori, Aaron and Dubey, Abhimanyu and Jauhri, Abhinav and Pandey, Abhinav and Kadian, Abhishek and Al-Dahle, Ahmad and Letman, Aiesha and Mathur, Akhil and Schelten, Alan and Vaughan, Alex and others},
  journal={arXiv preprint arXiv:2407.21783},
  year={2024}
}

@article{liu2019roberta,
  title={Roberta: A robustly optimized bert pretraining approach},
  author={Liu, Yinhan and Ott, Myle and Goyal, Naman and Du, Jingfei and Joshi, Mandar and Chen, Danqi and Levy, Omer and Lewis, Mike and Zettlemoyer, Luke and Stoyanov, Veselin},
  journal={arXiv preprint arXiv:1907.11692},
  year={2019}
}

@inproceedings{jiang-etal-2021-im,
    title = "{\textquotedblleft}{I}`m Not Mad{\textquotedblright}: Commonsense Implications of Negation and Contradiction",
    author = "Jiang, Liwei  and
      Bosselut, Antoine  and
      Bhagavatula, Chandra  and
      Choi, Yejin",
    editor = "Toutanova, Kristina  and
      Rumshisky, Anna  and
      Zettlemoyer, Luke  and
      Hakkani-Tur, Dilek  and
      Beltagy, Iz  and
      Bethard, Steven  and
      Cotterell, Ryan  and
      Chakraborty, Tanmoy  and
      Zhou, Yichao",
    booktitle = "Proceedings of the 2021 Conference of the North American Chapter of the Association for Computational Linguistics: Human Language Technologies",
    month = jun,
    year = "2021",
    address = "Online",
    publisher = "Association for Computational Linguistics",
    url = "https://aclanthology.org/2021.naacl-main.346/",
    doi = "10.18653/v1/2021.naacl-main.346",
    pages = "4380--4397"
}

@article{dettmers2023qlora,
  title={Qlora: Efficient finetuning of quantized llms},
  author={Dettmers, Tim and Pagnoni, Artidoro and Holtzman, Ari and Zettlemoyer, Luke},
  journal={Advances in neural information processing systems},
  volume={36},
  pages={10088--10115},
  year={2023}
}

@inproceedings{ravichander-etal-2022-condaqa,
    title = "{CONDAQA}: A Contrastive Reading Comprehension Dataset for Reasoning about Negation",
    author = "Ravichander, Abhilasha  and
      Gardner, Matt  and
      Marasovic, Ana",
    editor = "Goldberg, Yoav  and
      Kozareva, Zornitsa  and
      Zhang, Yue",
    booktitle = "Proceedings of the 2022 Conference on Empirical Methods in Natural Language Processing",
    month = dec,
    year = "2022",
    address = "Abu Dhabi, United Arab Emirates",
    publisher = "Association for Computational Linguistics",
    url = "https://aclanthology.org/2022.emnlp-main.598/",
    doi = "10.18653/v1/2022.emnlp-main.598",
    pages = "8729--8755"
}

@inproceedings{talmor-etal-2019-commonsenseqa,
    title = "{C}ommonsense{QA}: A Question Answering Challenge Targeting Commonsense Knowledge",
    author = "Talmor, Alon  and
      Herzig, Jonathan  and
      Lourie, Nicholas  and
      Berant, Jonathan",
    editor = "Burstein, Jill  and
      Doran, Christy  and
      Solorio, Thamar",
    booktitle = "Proceedings of the 2019 Conference of the North {A}merican Chapter of the Association for Computational Linguistics: Human Language Technologies, Volume 1 (Long and Short Papers)",
    month = jun,
    year = "2019",
    address = "Minneapolis, Minnesota",
    publisher = "Association for Computational Linguistics",
    url = "https://aclanthology.org/N19-1421/",
    doi = "10.18653/v1/N19-1421",
    pages = "4149--4158"
}

@article{guan-etal-2020-knowledge,
    title = "A Knowledge-Enhanced Pretraining Model for Commonsense Story Generation",
    author = "Guan, Jian  and
      Huang, Fei  and
      Zhao, Zhihao  and
      Zhu, Xiaoyan  and
      Huang, Minlie",
    editor = "Johnson, Mark  and
      Roark, Brian  and
      Nenkova, Ani",
    journal = "Transactions of the Association for Computational Linguistics",
    volume = "8",
    year = "2020",
    address = "Cambridge, MA",
    publisher = "MIT Press",
    url = "https://aclanthology.org/2020.tacl-1.7/",
    doi = "10.1162/tacl_a_00302",
    pages = "93--108"
}

@inproceedings{west-etal-2022-symbolic,
    title = "Symbolic Knowledge Distillation: from General Language Models to Commonsense Models",
    author = "West, Peter  and
      Bhagavatula, Chandra  and
      Hessel, Jack  and
      Hwang, Jena  and
      Jiang, Liwei  and
      Le Bras, Ronan  and
      Lu, Ximing  and
      Welleck, Sean  and
      Choi, Yejin",
    editor = "Carpuat, Marine  and
      de Marneffe, Marie-Catherine  and
      Meza Ruiz, Ivan Vladimir",
    booktitle = "Proceedings of the 2022 Conference of the North American Chapter of the Association for Computational Linguistics: Human Language Technologies",
    month = jul,
    year = "2022",
    address = "Seattle, United States",
    publisher = "Association for Computational Linguistics",
    url = "https://aclanthology.org/2022.naacl-main.341/",
    doi = "10.18653/v1/2022.naacl-main.341",
    pages = "4602--4625"
}

@inproceedings{bosselut-etal-2019-comet,
    title = "{COMET}: Commonsense Transformers for Automatic Knowledge Graph Construction",
    author = "Bosselut, Antoine  and
      Rashkin, Hannah  and
      Sap, Maarten  and
      Malaviya, Chaitanya  and
      Celikyilmaz, Asli  and
      Choi, Yejin",
    editor = "Korhonen, Anna  and
      Traum, David  and
      M{\`a}rquez, Llu{\'i}s",
    booktitle = "Proceedings of the 57th Annual Meeting of the Association for Computational Linguistics",
    month = jul,
    year = "2019",
    address = "Florence, Italy",
    publisher = "Association for Computational Linguistics",
    url = "https://aclanthology.org/P19-1470/",
    doi = "10.18653/v1/P19-1470",
    pages = "4762--4779"
}

@inproceedings{zhao-etal-2024-uncommonsense,
    title = "{UN}commonsense Reasoning: Abductive Reasoning about Uncommon Situations",
    author = "Zhao, Wenting  and
      Chiu, Justin  and
      Hwang, Jena  and
      Brahman, Faeze  and
      Hessel, Jack  and
      Choudhury, Sanjiban  and
      Choi, Yejin  and
      Li, Xiang  and
      Suhr, Alane",
    editor = "Duh, Kevin  and
      Gomez, Helena  and
      Bethard, Steven",
    booktitle = "Proceedings of the 2024 Conference of the North American Chapter of the Association for Computational Linguistics: Human Language Technologies (Volume 1: Long Papers)",
    month = jun,
    year = "2024",
    address = "Mexico City, Mexico",
    publisher = "Association for Computational Linguistics",
    url = "https://aclanthology.org/2024.naacl-long.469/",
    doi = "10.18653/v1/2024.naacl-long.469",
    pages = "8487--8505"
}

@article{li2024generation,
  title={From generation to judgment: Opportunities and challenges of llm-as-a-judge},
  author={Li, Dawei and Jiang, Bohan and Huang, Liangjie and Beigi, Alimohammad and Zhao, Chengshuai and Tan, Zhen and Bhattacharjee, Amrita and Jiang, Yuxuan and Chen, Canyu and Wu, Tianhao and others},
  journal={arXiv preprint arXiv:2411.16594},
  year={2024}
}

@inproceedings{nogueira-etal-2020-document,
    title = "Document Ranking with a Pretrained Sequence-to-Sequence Model",
    author = "Nogueira, Rodrigo  and
      Jiang, Zhiying  and
      Pradeep, Ronak  and
      Lin, Jimmy",
    editor = "Cohn, Trevor  and
      He, Yulan  and
      Liu, Yang",
    booktitle = "Findings of the Association for Computational Linguistics: EMNLP 2020",
    month = nov,
    year = "2020",
    address = "Online",
    publisher = "Association for Computational Linguistics",
    url = "https://aclanthology.org/2020.findings-emnlp.63/",
    doi = "10.18653/v1/2020.findings-emnlp.63",
    pages = "708--718"
}

@inproceedings{hossain-etal-2020-analysis,
    title = "An Analysis of Natural Language Inference Benchmarks through the Lens of Negation",
    author = "Hossain, Md Mosharaf  and
      Kovatchev, Venelin  and
      Dutta, Pranoy  and
      Kao, Tiffany  and
      Wei, Elizabeth  and
      Blanco, Eduardo",
    editor = "Webber, Bonnie  and
      Cohn, Trevor  and
      He, Yulan  and
      Liu, Yang",
    booktitle = "Proceedings of the 2020 Conference on Empirical Methods in Natural Language Processing (EMNLP)",
    month = nov,
    year = "2020",
    address = "Online",
    publisher = "Association for Computational Linguistics",
    url = "https://aclanthology.org/2020.emnlp-main.732/",
    doi = "10.18653/v1/2020.emnlp-main.732",
    pages = "9106--9118"
}

@inproceedings{weller-etal-2024-nevir,
    title = "{N}ev{IR}: Negation in Neural Information Retrieval",
    author = "Weller, Orion  and
      Lawrie, Dawn  and
      Van Durme, Benjamin",
    editor = "Graham, Yvette  and
      Purver, Matthew",
    booktitle = "Proceedings of the 18th Conference of the European Chapter of the Association for Computational Linguistics (Volume 1: Long Papers)",
    month = mar,
    year = "2024",
    address = "St. Julian{'}s, Malta",
    publisher = "Association for Computational Linguistics",
    url = "https://aclanthology.org/2024.eacl-long.139/",
    pages = "2274--2287"
}

@inproceedings{hosseini-etal-2021-understanding,
    title = "Understanding by Understanding Not: Modeling Negation in Language Models",
    author = "Hosseini, Arian  and
      Reddy, Siva  and
      Bahdanau, Dzmitry  and
      Hjelm, R Devon  and
      Sordoni, Alessandro  and
      Courville, Aaron",
    editor = "Toutanova, Kristina  and
      Rumshisky, Anna  and
      Zettlemoyer, Luke  and
      Hakkani-Tur, Dilek  and
      Beltagy, Iz  and
      Bethard, Steven  and
      Cotterell, Ryan  and
      Chakraborty, Tanmoy  and
      Zhou, Yichao",
    booktitle = "Proceedings of the 2021 Conference of the North American Chapter of the Association for Computational Linguistics: Human Language Technologies",
    month = jun,
    year = "2021",
    address = "Online",
    publisher = "Association for Computational Linguistics",
    url = "https://aclanthology.org/2021.naacl-main.102/",
    doi = "10.18653/v1/2021.naacl-main.102",
    pages = "1301--1312"
}

@inproceedings{rezaei-blanco-2025-making,
    title = "Making Language Models Robust Against Negation",
    author = "Rezaei, MohammadHossein  and
      Blanco, Eduardo",
    editor = "Chiruzzo, Luis  and
      Ritter, Alan  and
      Wang, Lu",
    booktitle = "Proceedings of the 2025 Conference of the Nations of the Americas Chapter of the Association for Computational Linguistics: Human Language Technologies (Volume 1: Long Papers)",
    month = apr,
    year = "2025",
    address = "Albuquerque, New Mexico",
    publisher = "Association for Computational Linguistics",
    url = "https://aclanthology.org/2025.naacl-long.413/",
    pages = "8123--8142",
    ISBN = "979-8-89176-189-6"
}

@inproceedings{lal-etal-2022-using,
    title = "Using Commonsense Knowledge to Answer Why-Questions",
    author = "Lal, Yash Kumar  and
      Tandon, Niket  and
      Aggarwal, Tanvi  and
      Liu, Horace  and
      Chambers, Nathanael  and
      Mooney, Raymond  and
      Balasubramanian, Niranjan",
    editor = "Goldberg, Yoav  and
      Kozareva, Zornitsa  and
      Zhang, Yue",
    booktitle = "Proceedings of the 2022 Conference on Empirical Methods in Natural Language Processing",
    month = dec,
    year = "2022",
    address = "Abu Dhabi, United Arab Emirates",
    publisher = "Association for Computational Linguistics",
    url = "https://aclanthology.org/2022.emnlp-main.79/",
    doi = "10.18653/v1/2022.emnlp-main.79",
    pages = "1204--1219"
}

@inproceedings{bowman-etal-2015-large,
    title = "A large annotated corpus for learning natural language inference",
    author = "Bowman, Samuel R.  and
      Angeli, Gabor  and
      Potts, Christopher  and
      Manning, Christopher D.",
    editor = "M{\`a}rquez, Llu{\'i}s  and
      Callison-Burch, Chris  and
      Su, Jian",
    booktitle = "Proceedings of the 2015 Conference on Empirical Methods in Natural Language Processing",
    month = sep,
    year = "2015",
    address = "Lisbon, Portugal",
    publisher = "Association for Computational Linguistics",
    url = "https://aclanthology.org/D15-1075/",
    doi = "10.18653/v1/D15-1075",
    pages = "632--642"
}

@article{brown2020language,
  title={Language models are few-shot learners},
  author={Brown, Tom and Mann, Benjamin and Ryder, Nick and Subbiah, Melanie and Kaplan, Jared D and Dhariwal, Prafulla and Neelakantan, Arvind and Shyam, Pranav and Sastry, Girish and Askell, Amanda and others},
  journal={Advances in neural information processing systems},
  volume={33},
  pages={1877--1901},
  year={2020}
}

@inproceedings{williams-etal-2018-broad,
    title = "A Broad-Coverage Challenge Corpus for Sentence Understanding through Inference",
    author = "Williams, Adina  and
      Nangia, Nikita  and
      Bowman, Samuel",
    editor = "Walker, Marilyn  and
      Ji, Heng  and
      Stent, Amanda",
    booktitle = "Proceedings of the 2018 Conference of the North {A}merican Chapter of the Association for Computational Linguistics: Human Language Technologies, Volume 1 (Long Papers)",
    month = jun,
    year = "2018",
    address = "New Orleans, Louisiana",
    publisher = "Association for Computational Linguistics",
    url = "https://aclanthology.org/N18-1101/",
    doi = "10.18653/v1/N18-1101",
    pages = "1112--1122"
}

@inproceedings{10.1007/11736790_9,
author = {Dagan, Ido and Glickman, Oren and Magnini, Bernardo},
title = {The PASCAL recognising textual entailment challenge},
year = {2005},
isbn = {3540334270},
publisher = {Springer-Verlag},
address = {Berlin, Heidelberg},
url = {https://doi.org/10.1007/11736790_9},
doi = {10.1007/11736790_9},
booktitle = {Proceedings of the First International Conference on Machine Learning Challenges: Evaluating Predictive Uncertainty Visual Object Classification, and Recognizing Textual Entailment},
pages = {177–190},
numpages = {14},
location = {Southampton, UK},
series = {MLCW'05}
}

@article{10.1162/coli.07-034-R2,
    author = {Artstein, Ron and Poesio, Massimo},
    title = {Inter-Coder Agreement for Computational Linguistics},
    journal = {Computational Linguistics},
    volume = {34},
    number = {4},
    pages = {555-596},
    year = {2008},
    month = {12},
    issn = {0891-2017},
    doi = {10.1162/coli.07-034-R2},
    url = {https://doi.org/10.1162/coli.07-034-R2},
    eprint = {https://direct.mit.edu/coli/article-pdf/34/4/555/1808947/coli.07-034-r2.pdf},
}

@article{mcnemar1947note,
  title={Note on the sampling error of the difference between correlated proportions or percentages},
  author={McNemar, Quinn},
  journal={Psychometrika},
  volume={12},
  number={2},
  pages={153--157},
  year={1947},
  publisher={Springer-Verlag}
}

@inproceedings{dobreva-keller-2021-investigating,
    title = "Investigating Negation in Pre-trained Vision-and-language Models",
    author = "Dobreva, Radina  and
      Keller, Frank",
    editor = "Bastings, Jasmijn  and
      Belinkov, Yonatan  and
      Dupoux, Emmanuel  and
      Giulianelli, Mario  and
      Hupkes, Dieuwke  and
      Pinter, Yuval  and
      Sajjad, Hassan",
    booktitle = "Proceedings of the Fourth BlackboxNLP Workshop on Analyzing and Interpreting Neural Networks for NLP",
    month = nov,
    year = "2021",
    address = "Punta Cana, Dominican Republic",
    publisher = "Association for Computational Linguistics",
    url = "https://aclanthology.org/2021.blackboxnlp-1.27/",
    doi = "10.18653/v1/2021.blackboxnlp-1.27",
    pages = "350--362"
}

@inproceedings{devlin-etal-2019-bert,
    title = "{BERT}: Pre-training of Deep Bidirectional Transformers for Language Understanding",
    author = "Devlin, Jacob  and
      Chang, Ming-Wei  and
      Lee, Kenton  and
      Toutanova, Kristina",
    editor = "Burstein, Jill  and
      Doran, Christy  and
      Solorio, Thamar",
    booktitle = "Proceedings of the 2019 Conference of the North {A}merican Chapter of the Association for Computational Linguistics: Human Language Technologies, Volume 1 (Long and Short Papers)",
    month = jun,
    year = "2019",
    address = "Minneapolis, Minnesota",
    publisher = "Association for Computational Linguistics",
    url = "https://aclanthology.org/N19-1423/",
    doi = "10.18653/v1/N19-1423",
    pages = "4171--4186"
}

@inproceedings{jang-etal-2022-beyond,
    title = "Beyond Distributional Hypothesis: Let Language Models Learn Meaning-Text Correspondence",
    author = "Jang, Myeongjun  and
      Mtumbuka, Frank  and
      Lukasiewicz, Thomas",
    editor = "Carpuat, Marine  and
      de Marneffe, Marie-Catherine  and
      Meza Ruiz, Ivan Vladimir",
    booktitle = "Findings of the Association for Computational Linguistics: NAACL 2022",
    month = jul,
    year = "2022",
    address = "Seattle, United States",
    publisher = "Association for Computational Linguistics",
    url = "https://aclanthology.org/2022.findings-naacl.156/",
    doi = "10.18653/v1/2022.findings-naacl.156",
    pages = "2030--2042"
}

@inproceedings{cer-etal-2017-semeval,
    title = "{S}em{E}val-2017 Task 1: Semantic Textual Similarity Multilingual and Crosslingual Focused Evaluation",
    author = "Cer, Daniel  and
      Diab, Mona  and
      Agirre, Eneko  and
      Lopez-Gazpio, I{\~n}igo  and
      Specia, Lucia",
    editor = "Bethard, Steven  and
      Carpuat, Marine  and
      Apidianaki, Marianna  and
      Mohammad, Saif M.  and
      Cer, Daniel  and
      Jurgens, David",
    booktitle = "Proceedings of the 11th International Workshop on Semantic Evaluation ({S}em{E}val-2017)",
    month = aug,
    year = "2017",
    address = "Vancouver, Canada",
    publisher = "Association for Computational Linguistics",
    url = "https://aclanthology.org/S17-2001/",
    doi = "10.18653/v1/S17-2001",
    pages = "1--14"
}

\appendix
\section{Details for Creating and Annotating the Benchmark}
\label{app:annotation_details}

We create the benchmark by sampling 200 triples per relation (9 relations in total) from \textsc{Atomic}'s test split.
The benchmark includes 7,200 triples augmented with three types of negation per original triple (negating the \textit{if} event, \textit{then} event, or both).
We ask two annotators, one graduate student and one with a PhD degree, to validate each triple using three labels: \emph{Valid}, \emph{Invalid}, and \emph{Ambiguous}.
Table~\ref{t:annotate_instruction} shows the instructions we provide to annotators.

\section{Details for Validating Commonsense Knowledge Triples}
\label{app:validating_triples}
We train a task-specific LLM to validate augmented commonsense triples.
The model is trained with the same amount of data regardless of the corpus source.
Specifically, we sample 1,800 triples per label from the training split of each source corpus (in total 5,400).
For \emph{Valid} and \emph{Ambiguous} triples, we sample 100 triples per relation per corpus if the sources are \textsc{Atomic} and \textsc{Anion}; 
otherwise, we sample 200 triples per relation from the single corpus.

For \emph{Invalid} triples, we use GPT-4o to generate \emph{then} events given sampled \emph{if} events from \textsc{Atomic} and \textsc{Anion} to construct \emph{Invalid} triples. 
Table~\ref{t:invalid_prompt} provides the prompt.

We train the task-specific LLM judge based on Llama 3.1 Instruct 8B and 70B using QLoRA~\cite{dettmers2023qlora} with 4-bit quantization,
which are hosted locally on two H100 GPUs with a total of 160 GB memory. 
Llama 3.1 8B is trained for 5 epochs with a learning rate of 5e-6 with 1 hour of training time,
while Llama 3.1 70B is trained for 1 epoch with a learning rate of 2e-5 with 16 hours of training time.

\begin{table}[t!]
  \centering
  \small
  \begin{tabular}{p{7cm} }
    \toprule

    Given a triple <\textit{A}, \textit{R}, \textit{B}> and its three negated variations 
    <\textit{$\neg$A}, \textit{R}, \textit{B}>, <\textit{A}, \textit{R}, \textit{$\neg$B}>, and <\textit{$\neg$A}, \textit{R}, \textit{$\neg$B}>,
    the annotation task is to determine if each triple aligns with real-world commonsense knowledge. 
    We use label \emph{Valid} to represent a triple that always aligns with real-world commonsense knowledge;
    label \emph{Invalid} to represent a triple that always conflicts with real-world commonsense knowledge;
    and label \emph{Ambiguous} to represent a triple for two cases: (1) the interpretation of the triple is ambiguous, 
    meaning it can either align or conflict with real-world commonsense knowledge, 
    or (2) the \emph{if} event does not relate to the \emph{then} event by the relation.\\\\
    Below are some examples.\\
    \emph{Valid}: If PersonX takes a picture, then PersonX wants to look at the picture.\\
    \emph{Invalid}: If PersonX takes a picture, then PersonX wants to not look at the picture.\\
    \emph{Ambiguous}: If PersonX takes a picture, then PersonX wants to take a nap. \\
    \emph{Ambiguous}: If PersonX opens the window, then PersonX wants to breathe. \\
    
    \bottomrule
    
\end{tabular}
  \caption{
    Annotation instructions for the benchmark.
    }
  \label{t:annotate_instruction}

\end{table}

\begin{table}[h]
  \centering
  \small
  \resizebox{0.48\textwidth}{!}{%
  \begin{tabular}{p{7.6cm} }
    \toprule
    You are an expert in commonsense reasoning and knowledge generation. 
    Your task is to generate a then event complementing the given if event so that the if-then statement conflicts with commonsense knowledge.\\
    These invalid statements should be clearly wrong or illogical based on everyday commonsense knowledge.\\
    \\
    Given the following incomplete if-then statement: \\
    Statement: If {event}, then {relation} ... \\ 
    Generate the then event within a phrase. \\
    \bottomrule
    
\end{tabular}
  }
  \caption{
    Prompt to generate if-then statements that conflict with commonsense knowledge. They are considered \textit{Invalid} triples for the validation task and downstream task.
    }
  \label{t:invalid_prompt}
\end{table}

\begin{table}[t!]
  \centering
  \small
  \begin{tabular}{l l}
    \toprule

    Relation & Verbalization \\
    \midrule
    oEffect  & the effect of \{object\} is \\
    oReact   & the reaction of \{object\} is \\
    oWant    & \{object\} want \\
    xAttr    & the attribute of PersonX is \\
    xEffect  & the effect of PersonX is \\
    xIntent  & the intention of PersonX is \\
    xNeed    & PersonX needs \\
    xReact   & the reaction of PersonX is \\
    xWant    & PersonX wants \\
    \bottomrule
    
\end{tabular}
  \caption{
    The mapping from commonsense knowledge triples with nine \emph{if-then} relations to natural language statements.
    \{object\} indicates the object in the \emph{if} event.
    }
  \label{t:verbalize_mapping}

\end{table}

Table~\ref{t:verbalize_mapping} lists the mapping we use to convert commonsense knowledge triples to natural language \emph{if-then} statements.
The models are trained and evaluated with the \emph{if-then} statements instead of the triples.

\begin{table}[t!]
  \centering
  \small
  \begin{tabular}{p{7.2cm} }
    \toprule
    You are a helpful assistant. In this task, you are expected to write answers to questions involving reasoning about negation.\\
    The answer to the question should be "yes", "no", "don't know", or a phrase in the passage. Questions can only have one correct answer.\\ 
    Only output [YES], [NO], [DON'T KNOW] or a short phrase in the passage.\\
    \\
    \{4 exemplars sampled from the few-shot learning split of CondaQA\}\\
    \bottomrule
    
\end{tabular}
  \caption{
    4-shot prompts to evaluate LLMs with CondaQA. We randomly sample 4 exemplars from the few-shot learning split provided by~\citet{ravichander-etal-2022-condaqa}.
    }
  \label{t:condaqa_prompt}
\end{table}

\begin{table}[t!]
  \centering
  \small
  \begin{tabular}{lrrr}
  \toprule
  & Learning Rate & Batch Size & Epochs \\
  \midrule
  RTE-neg & 2e-5 & 8 & 15 \\
  SNLI-neg & 2e-5 & 32 & 3 \\
  MNLI-neg & 2e-5 & 64 & 4 \\
  \bottomrule
  \end{tabular}
  \caption{   
      Hyperparameters for fully-supervised fine-tuning with three NLI benchmarks: RTE-neg, SNLI-neg, and MNLI-neg.}
  \label{t:nli_hyperparameters}
\end{table}

\subsection{Additional Results on Validating Commonsense Knowledge Triples}
\label{app:additional_validating_results}

Table~\ref{t:validation_results_full} reports additional results for validating commonsense triples with negation,
including F1 for each relation and overall Accuracy (Acc).
Among individual relations, xIntent and xAttr consistently yield higher F1 across models, while oEffect and oReact are more challenging.
Fine-tuning with \textsc{Atomic} generally outperforms \textsc{Anion} as the source of \textit{Valid} and \textit{Ambiguous} training data.

\begin{table*}
    \centering
    \small
    \setlength{\tabcolsep}{4pt}
\begin{tabular}{l rrrrrrrrr rr}
    \toprule
    & \multirow{2}{*}{oEffect} & \multirow{2}{*}{oReact}  & \multirow{2}{*}{oWant} & \multirow{2}{*}{xAttr}	
    & \multirow{2}{*}{xEffect} & \multirow{2}{*}{xIntent} &	\multirow{2}{*}{xNeed} & \multirow{2}{*}{xReact} 
    &\multirow{2}{*}{xWant}  & \multicolumn{2}{c}{All Relations} \\
    \cmidrule(lr){11-12}
    & & & & & & & & & &F1 & Acc\\
    \midrule
    Few-shot learning \\ 
    ~~~Llama 3.1 8B      & 0.34 & 0.38 & 0.34 & 0.39 & 0.35 & 0.50 & 0.31 & 0.34 & 0.36 & 0.37 & 0.44\\
    ~~~Llama 3.1 70B     & 0.48 & 0.45 & 0.53 & 0.56 & 0.56 & 0.56 & 0.42 & 0.52 & 0.57 & 0.53 & 0.53 \\ 
    ~~~GPT-4o            & 0.52 & 0.46 & 0.54 & 0.54 & 0.44 & 0.57 & 0.51 & 0.54 & 0.57 & 0.52 & 0.54\\ 
    ~~~Claude Sonnet 4   & 0.40 & 0.37 & 0.57 & 0.66 & 0.59 & 0.61 & 0.56 & 0.54 & 0.63 & 0.56 & 0.56\\

    \addlinespace
    
    Fine-tuning \\
    ~~~Llama 3.1 8B with\\
    ~~~~~~\textsc{Atomic}            & 0.51 & 0.57 & 0.49 & 0.63 & 0.53 & 0.62 & 0.50 & 0.53 & 0.52 & 0.55 & 0.56\\ 
    ~~~~~~\textsc{Anion}             & 0.52 & 0.56 & 0.50 & 0.59 & 0.51 & 0.59 & 0.45 & 0.56 & 0.50 & 0.53 & 0.55 \\ 
    ~~~~~~\textsc{Atomic} + \textsc{Anion}    & 0.38 & 0.44 & 0.46 & 0.57 & 0.47 & 0.56 & 0.48 & 0.51 & 0.47 & 0.48 & 0.51 \\
    \addlinespace
    ~~~Llama 3.1 70B with\\
    ~~~~~~\textsc{Atomic}            & 0.59 & 0.60 & 0.61 & 0.72 & 0.62 & 0.69 & 0.59 & 0.62 & 0.63 & 0.63 & 0.64 \\ 
    ~~~~~~\textsc{Anion}             & 0.54 & 0.54 & 0.56 & 0.68 & 0.51 & 0.60 & 0.43 & 0.50 & 0.55 & 0.55 & 0.56 \\ 
    ~~~~~~\textsc{Atomic} + \textsc{Anion}    & 0.56 & 0.58 & 0.61 & 0.71 & 0.60 & 0.66 & 0.57 & 0.62 & 0.63 & 0.62 & 0.62\\
     \bottomrule
    
    \end{tabular}

    \caption{
         Complete results of validating commonsense triples with negation.
         We report F1 for each relation and overall, along with overall Accuracy (Acc).
         We further report the complete fine-tuning results across variants that differ in whether \textit{Valid} and \textit{Ambiguous} instances come from \textsc{Atomic}, \textsc{Anion}, or both.
         The results complement Table~\ref{t:validation_results} in the main paper.}
      \label{t:validation_results_full}

  \end{table*}

\section{Experimental Details for Downstream Tasks}
\label{app:evaluation_details}

We train three models with either existing corpora or our commonsense knowledge corpora.
RoBERTa-large~\cite{liu2019roberta} is further fine-tuned with the training split of the specific downstream task,
using a batch size of 128, a learning rate of 1e-6, and early stopping with patience of 3 epochs and a maximum of 5 epochs.

For the two LLMs (Llama 3.1 8B~\cite{grattafiori2024llama} and Qwen2 7B~\cite{yang2024qwen2technicalreport}), 
we adopt a standard instruction fine-tuning paradigm, 
where the training input is formatted as: [instruction, verbalized commonsense triple, output label], 
and the model is trained using the causal language modeling objective.
Specifically, they are trained for 2 epochs using a batch size of 16; Llama 3.1 8B uses a learning rate of 5e-6 and Qwen2 7B uses 2e-5.
We also use QLoRA~\cite{dettmers2023qlora} with 4-bit quantization for both models.
Each model takes approximately 4 hours to train on one H100 GPU with 80 GB memory.
They are further evaluated using few-shot prompting.
GPT-4o is called via OpenAI's API and Claude Sonnet 4 is called via the AWS Bedrock API.

\subsection{Experimental Details for CondaQA}

For fully-supervised evaluation, we train LLMs with CondaQA's training split using a batch size of 8 and a learning rate of 1e-5.
We use early stopping with a patience of 3 epochs and a maximum of 5 epochs.

Table~\ref{t:condaqa_prompt} reports the 4-shot prompts for evaluating LLMs with CondaQA.
We reuse the prompts with minimal edits and sample 4 exemplars from the few-shot learning split provided by~\citet{ravichander-etal-2022-condaqa}.

\subsection{Experimental Details for NLI Benchmarks}

Table~\ref{t:nli_hyperparameters} reports the hyperparameters for fully-supervised fine-tuning LLMs with NLI datasets.

Table~\ref{t:rte_prompt} reports
the 4-shot prompts used to evaluate LLMs with RTE-neg dataset, and Table~\ref{t:nli_prompt} reports
the 4-shot prompts for evaluating with MNLI-neg and SNLI-neg datasets. 
All exemplars are sampled from their training split.

\begin{table*}[t!]
  \centering
  \small
  \begin{tabular}{p{15.4cm} }
    \toprule

    You are a helpful assistant. You are given a pair of sentences: a premise and a hypothesis. Your task is to determine the relationship between the two sentences.\\
    \#\# [entailment]: The premise guarantees the truth of the hypothesis.\\
    \#\# [not\_entailment]: The premise does not guarantee the truth of the hypothesis. \\
    Format your answer by only outputting [entailment] or [not\_entailment].\\\\

    \#\# Premise: Edward VIII became King in January of 1936 and abdicated in December.\\
    \#\# Hypothesis: King Edward VIII abdicated in December 1936.\\
    \#\# Response: [entailment]\\\\
    
    \#\# Premise: Oil prices fall back as Yukos oil threat lifted.\\
    \#\# Hypothesis: Oil prices rise.\\
    \#\# Response: [not\_entailment]\\\\
    
    \#\# Premise: World Bank programs have been heavily criticized for many years for resulting in poverty.\\
    \#\# Hypothesis: The World Bank is criticized for its activities.\\
    \#\# Response: [entailment]\\\\

    \#\# Premise: The cost of the consumer of the United States fell in June.\\
    \#\# Hypothesis: U.S. consumer spending dived in June.\\
    \#\# Response: [not\_entailment]\\\\
    
    \bottomrule
    
\end{tabular}
\caption{
  4-shot prompts for evaluating LLMs with RTE-neg dataset. The exemplars are chosen from the training split.
  }
\label{t:rte_prompt}
\end{table*}

\begin{table*}[t!]
  \centering
  \small
\begin{tabular}{p{15.4cm}}
  \toprule

  You are a helpful assistant. You are given a pair of sentences: a premise and a hypothesis. Your task is to determine the relationship between the two sentences.\\
  \#\# [entailment]: The hypothesis is definitely true given the premise.\\
  \#\# [contradiction]: The hypothesis is definitely false given the premise.\\
  \#\# [neutral]: It is not possible to determine whether the hypothesis is true or false just from the premise. \\
  Format your answer by only outputting [entailment], [contradiction], or [neutral].\\\\

  \#\# Premise: One of our number will carry out your instructions minutely.\\
  \#\# Hypothesis: A member of my team will execute your orders with immense precision.\\ 
  \#\# Response: [entailment]\\\\

  \#\# Premise: Fun for adults and children.\\
  \#\# Hypothesis: Fun for only children.\\
  \#\# Response: [contradiction]\\\\
  
  \#\# Premise: He turned and smiled at Vrenna.\\
  \#\# Hypothesis: He smiled at Vrenna who was walking slowly behind him with her mother.\\
  \#\# Response: [neutral]\\\\

  \#\# Premise: The famous tenements (or lands) began to be built.\\
  \#\# Hypothesis: The land remained deserted.\\
  \#\# Response: [contradiction]\\\\
  
  \bottomrule
  
\end{tabular}

  \caption{
    4-shot prompts for evaluating LLMs with SNLI-neg and MNLI-neg datasets. The exemplars are chosen from the training split.
    }
  \label{t:nli_prompt}
\end{table*}

\subsection{Experimental Details for NevIR}

As NevIR does not provide any training data,
following~\citet{weller-etal-2024-nevir}, we perform fully-supervised fine-tuning with LLMs on STS-B~\cite{cer-etal-2017-semeval}
using a learning rate of 2e-5 and a batch size of 32 for 4 epochs.

Table~\ref{t:nevir_prompt} reports the 4-shot prompts for evaluating with NevIR.

\begin{table*}[t!]
  \centering
  \small
  \begin{tabular}{p{15.4cm} }
    \toprule

    You are a helpful assistant. You are given a query and two documents. 
    Your task is to choose the document that has the answer for the query.\\
    Output [Doc1] if the first document has the answer for the query, or [Doc2] if the second document has the answer.\\\\

    \#\# Query: Which mayor did more vetoing than anticipated?\\
    \#\# Doc1: In his first year as mayor, Medill received very little legislative resistance from the Chicago City Council. While he vetoed what was an unprecedented eleven City Council ordinances that year, most narrowly were involved with specific financial practices considered wasteful and none of the vetoes were overridden. He used his new powers to appoint the members of the newly constituted Chicago Board of Education and the commissioners of its constituted public library. His appointments were approved unanimously by the City Council.\\
    \#\# Doc2: In his first year as mayor, Medill received very little legislative resistance from the Chicago City Council. While some expected an unprecedented number of vetoes, in actuality he only vetoed eleven City Council ordinances that year, and most of those were narrowly involved with specific financial practices he considered wasteful and none of the vetoes were overridden. He used his new powers to appoint the members of the newly constituted Chicago Board of Education and the commissioners of its constituted public library. His appointments were approved unanimously by the City Council.\\
    \#\# Response: [Doc1]\\\\
    
    \#\# Query: Which mayor did less vetoing than anticipated?\\
    \#\# Doc1: In his first year as mayor, Medill received very little legislative resistance from the Chicago City Council. While he vetoed what was an unprecedented eleven City Council ordinances that year, most narrowly were involved with specific financial practices considered wasteful and none of the vetoes were overridden. He used his new powers to appoint the members of the newly constituted Chicago Board of Education and the commissioners of its constituted public library. His appointments were approved unanimously by the City Council.\\
    \#\# Doc2: In his first year as mayor, Medill received very little legislative resistance from the Chicago City Council. While some expected an unprecedented number of vetoes, in actuality he only vetoed eleven City Council ordinances that year, and most of those were narrowly involved with specific financial practices he considered wasteful and none of the vetoes were overridden. He used his new powers to appoint the members of the newly constituted Chicago Board of Education and the commissioners of its constituted public library. His appointments were approved unanimously by the City Council.\\
    \#\# Response: [Doc2]\\\\
    
    \#\# Query: Which Swiss cantons do not have official churches?\\
    \#\# Doc1: Switzerland has no official state religion, though most of the cantons (except Geneva and Neuchâtel) recognise official churches, which are either the Roman Catholic Church or the Swiss Reformed Church. These churches, and in some cantons also the Old Catholic Church and Jewish congregations, are financed by official taxation of adherents.\\
    \#\# Doc2: Switzerland has no official state religion, though most of the cantons (except Neuchâtel) recognise official churches, which are either the Roman Catholic Church or the Swiss Reformed Church. These churches, and in some cantons also the Old Catholic Church and Jewish congregations, are financed by official taxation of adherents.\\
    \#\# Response: [Doc1]\\\\

    \#\# Query: Which Swiss canton does not have official churches?\\
    \#\# Doc1: Switzerland has no official state religion, though most of the cantons (except Geneva and Neuchâtel) recognise official churches, which are either the Roman Catholic Church or the Swiss Reformed Church. These churches, and in some cantons also the Old Catholic Church and Jewish congregations, are financed by official taxation of adherents.\\
    \#\# Doc2: Switzerland has no official state religion, though most of the cantons (except Neuchâtel) recognise official churches, which are either the Roman Catholic Church or the Swiss Reformed Church. These churches, and in some cantons also the Old Catholic Church and Jewish congregations, are financed by official taxation of adherents.\\
    \#\# Response: [Doc2]\\
    
    \bottomrule
    
\end{tabular}
  \caption{
    4-shot prompts to evaluate LLMs with NevIR.
    }
  \label{t:nevir_prompt}
\end{table*}

\section{Additional Results on Downstream Tasks}
\label{app:additional_results}

\subsection{Complete Results for Downstream Tasks}
\label{app:full_results}

Table~\ref{t:condaqa_evaluation} and Table~\ref{t:nli_nevIR_evaluation}
report only the best results among three training corpora configurations using either $\neg$\textsc{Atomic}, $\neg$\textsc{Anion}, or both.
Table~\ref{t:condaqa_evaluation_full} and Table~\ref{t:nli_nevIR_evaluation_full} provide complete results for all three configurations.

\begin{table*}[t]
    \centering
    \small
    \begin{tabular}{l r r@{ }l rrrr}
\toprule
                                                              &            &          &                  & \multicolumn{4}{c}{Group Consistency} \\
\cmidrule(lr){5-8}
                                                              & \# Params. & Accuracy & ($\Delta$\%)     &  All &  Par. &  Sco. &  Aff. \\
\midrule
\textbf{Fully supervised}                                     \\
~~~UnifiedQA-v2-large~\cite{ravichander-etal-2022-condaqa}    & 770M       &     66.7 &&                   30.2 &  64.0 &  43.7 &  46.5 \\
~~~RoBERTa-large                                              & 355M       &     64.9 &&                   29.6 &  61.9 &  41.4 &  45.8 \\
~~~~~~Further pre-trained with                                \\
~~~~~~~~~Original commonsense triples from                    \\
~~~~~~~~~~~~\textsc{Atomic} + \textsc{Anion}                  &&                 67.0 & (+3.2)          & 32.9 &  65.6 &  46.3 &  48.1 \\
~~~~~~~~~Negated commonsense triples from                     \\
~~~~~~~~~~~~$\neg$\textsc{Atomic}                             &&                 67.7 & (+4.3)          & 32.8 &  66.1 &  46.3 &  48.6 \\
~~~~~~~~~~~~$\neg$\textsc{Anion}                              &&                 67.9 & (+4.6)$^\ast$   & 33.3 & 66.0 & 45.8 & 50.1 \\
~~~~~~~~~~~~$\neg$\textsc{Atomic} + $\neg$\textsc{Anion}      &&        \textbf{68.5} & (+5.5)$^\ast$   & \textbf{34.3} & \textbf{66.4} & \textbf{47.6} & \textbf{50.1} \\
\midrule

\textbf{In-context learning}                                            \\

~~~\textbf{Zero-shot}                                            \\
~~~~~~OpenAI o1                                                  & ---        &     65.3 &&                   24.9 &  67.4 &  43.8 & 38.6 \\
~~~\textbf{Few-shot}                                             \\
~~~~~~InstructGPT + COT~\cite{ravichander-etal-2022-condaqa}     & ---        &     66.3 &&                   27.3 &  64.2 &  45.1 &  44.9 \\
~~~~~~GPT-4o                                                     & ---        &     72.9 &&                   34.4 &  78.7 &  52.6 &  47.9 \\
~~~~~~Llama 3.1 70B                                              & 70B        &     77.5 &&                   43.3 &  83.7 &  61.8 &  54.9 \\
~~~~~~Llama 3.1 8B                                               & 8B         &     68.7 &&                   31.5 &  69.0 &  48.1 &  44.4 \\
~~~~~~~~~Further pre-trained with                                \\
~~~~~~~~~~~~Original commonsense triples from                    \\
~~~~~~~~~~~~~~~\textsc{Atomic} + \textsc{Anion}                  &&                 67.6 & (-1.6)          & 27.5 &  69.5 &  44.3 &  41.4 \\
~~~~~~~~~~~~Negated commonsense triples from                     \\
~~~~~~~~~~~~~~~$\neg$\textsc{Atomic}                             &&                 70.6 & (+2.8)          & 32.8 &  71.8  &  48.6 &  45.0 \\
~~~~~~~~~~~~~~~$\neg$\textsc{Anion}                              &&                 71.3 & (+3.8)$^\ast$   & 33.4 &  72.5  & \textbf{49.6} & 45.7 \\
~~~~~~~~~~~~~~~$\neg$\textsc{Atomic} + $\neg$\textsc{Anion}      &&       \textbf{71.5}  & (+4.1)$^\ast$   & \textbf{33.5} & \textbf{72.9} & 48.7 & \textbf{46.4} \\
\addlinespace
~~~Qwen2 7B                                                   & 7B         &     65.1 &&                   24.9 &  65.9 &  39.7 &  39.2 \\
~~~~~~Further pre-trained with                                \\
~~~~~~~~~Original commonsense triples from                    \\ 
~~~~~~~~~~~~\textsc{Atomic} + \textsc{Anion}                  &&                 64.1 & (-1.5)           & 22.8 &  65.2 &  38.4 &  37.7 \\
~~~~~~~~~Negated commonsense triples from                     \\
~~~~~~~~~~~~$\neg$\textsc{Atomic}                             &&       \textbf{69.7}  & (+7.1)$^\ast$   & \textbf{32.7} & \textbf{71.2} & \textbf{46.3} & \textbf{45.5} \\
~~~~~~~~~~~~$\neg$\textsc{Anion}                              &&                 69.2 & (+6.2)$^\ast$   & 29.9 &  68.8 &  45.4 &  44.1 \\
~~~~~~~~~~~~$\neg$\textsc{Atomic} + $\neg$\textsc{Anion}      &&                 66.6 & (+2.3)          & 28.7 & 68.8 & 43.7 & 40.8 \\
\bottomrule
\end{tabular}
    \caption{
         Complete results evaluating CondaQA 
      with two settings: fully supervised (top) and in-context learning (bottom). 
      We experiment with three training configurations on our corpora.  
      The best result for each model is in bold.
      Delta ($\Delta$) indicates the percent change in accuracy compared to the off-the-shelf model.
      An asterisk $^\ast$ indicates a statistically significant
      improvement (McNemar’s test~\cite{mcnemar1947note}, $p < 0.05$) over both the off-the-shelf model and the model trained with existing corpora.
     }
      \label{t:condaqa_evaluation_full}

  \end{table*}

  \begin{table*}[t]
    \centering
    \small
    \begin{tabular}{l r r@{}l r@{}l r@{}l r@{}l}
\toprule
                                                       & \multirow{3}{*}{\# Params.} & \multicolumn{6}{c}{NLI with Negation}       & \multirow{3}{*}{NevIR} \\
\cmidrule(lr){3-8}
                                                       &      &          RTE-Neg &&         SNLI-Neg &&         MNLI-Neg          \\
\midrule
\textbf{Fully supervised}                              \\
~~~BERTNOT~\cite{hosseini-etal-2021-understanding}     & 110M &             74.5 &&             46.0 &&             60.9 &&   --- \\
~~~RoBERTa-large-NSP~\cite{rezaei-blanco-2025-making}  & 355M &             87.2 &&             56.5 &&             69.9 &&   --- \\
~~~stsb-roberta-large~\cite{weller-etal-2024-nevir}    & 355M &              --- &&              --- &&              --- &&  24.9 \\
~~~MonoT5 3B~\cite{nogueira-etal-2020-document}        & 3B   &              --- &&              --- &&              --- &&  50.6 \\
~~~RoBERTa-large                                       & 355M &             84.7 &&             56.0 &&             69.9 &&  24.5 \\
~~~~~~Further pre-trained with                         \\
~~~~~~~~~Original commonsense triples from             \\
~~~~~~~~~~~~\textsc{Atomic} + \textsc{Anion}             &&                 86.2 &&             56.5 &&             69.4 &&  29.1 \\
~~~~~~~~~Negated commonsense triples from              \\
~~~~~~~~~~~~$\neg$\textsc{Atomic}                               &&                 81.5 &&             56.6 &&   \textbf{70.9}  &&  29.8 \\
~~~~~~~~~~~~$\neg$\textsc{Anion}                                &&                 85.2 &&             57.1 &&             69.3 &&  30.5 \\
~~~~~~~~~~~~$\neg$\textsc{Atomic} + $\neg$\textsc{Anion}        && \textbf{88.1} &$^\ast$&   \textbf{58.3}  &&             69.7 && \textbf{34.3} &$^\ast$ \\
\midrule
\textbf{Zero-shot}                                     \\
~~~OpenAI o1                                           & ---  &             87.5 &&             75.9 &&             74.6 &&  59.7 \\
\textbf{Few-shot}                                      \\
~~~GPT-4o                                              & ---  &             86.9 &&             74.8 &&             75.0 &&  61.7 \\
~~~Llama 3.1 70B                                       & 70B  &             78.9 &&             69.1 &&             65.9 &&  58.8 \\
~~~Llama 3.1 8B                                        & 8B   &             60.0 &&             54.4 &&             47.0 &&  30.6 \\
~~~~~~Further pre-trained with                         \\
~~~~~~~~~Original commonsense triples from             \\
~~~~~~~~~~~~\textsc{Atomic} + \textsc{Anion}             &&                 65.3 &&             57.5 &&             51.1 &&  37.9 \\
~~~~~~~~~Negated commonsense triples from              \\
~~~~~~~~~~~~$\neg$\textsc{Atomic}                        &&                 81.2 &$^\ast$&  \textbf{68.5} &$^\ast$&        63.8 &$^\ast$&  38.1 & \\
~~~~~~~~~~~~$\neg$\textsc{Anion}                         &&        \textbf{81.3} &$^\ast$&          68.2  &$^\ast$&\textbf{63.9}&$^\ast$& \textbf{42.2}&$^\ast$ \\
~~~~~~~~~~~~$\neg$\textsc{Atomic} + $\neg$\textsc{Anion} &&                 81.3 &$^\ast$& 67.9  &$^\ast$&        63.7 &$^\ast$&  38.2 & \\
\addlinespace
~~~Qwen2 7B                                            & 7B   &             71.7 &&             60.7 &&             59.5 &&  29.8 \\
~~~~~~Further pre-trained with                         \\
~~~~~~~~~Original commonsense triples from             \\
~~~~~~~~~~~~\textsc{Atomic} + \textsc{Anion}           &&                   78.3 &&             66.5 &&             63.0 &&  33.8 \\
~~~~~~~~~Negated commonsense triples from              \\
~~~~~~~~~~~~$\neg$\textsc{Atomic}                      &&                 \textbf{82.3} &$^\ast$&     \textbf{72.4} &$^\ast$&  \textbf{67.5} &$^\ast$&  \textbf{39.8} &$^\ast$ \\
~~~~~~~~~~~~$\neg$\textsc{Anion}                       &&                  78.9         &&             66.1         &&              62.5     &&          36.3        \\
~~~~~~~~~~~~$\neg$\textsc{Atomic} + $\neg$\textsc{Anion}&&                 82.1         &$^\ast$&      72.2         &$^\ast$&       67.5     &$^\ast$&   38.5 &$^\ast$ \\
\bottomrule
\end{tabular}
    \caption{
         Complete results evaluating three NLI benchmarks with negation cues (accuracy) and NevIR benchmark (pairwise accuracy) in two settings: 
      (1) fully supervised (top) and (2) in-context learning (bottom). We experiment with three training configurations on our corpora.  
      The best result for each model is in bold.
      An asterisk $^\ast$ indicates a statistically significant improvement over both the off-the-shelf model and the model trained with existing corpora.  }
        \label{t:nli_nevIR_evaluation_full}

  \end{table*}
\subsection{Evaluation on CommonsenseQA}
\label{app:commonsenseqa}

We evaluate off-the-shelf and pre-trained models on CommonsenseQA~\cite{talmor-etal-2019-commonsenseqa}, 
a standard commonsense reasoning benchmark without negation, to verify that pre-training on our negated corpora does not degrade general commonsense reasoning ability.
Table~\ref{t:commonsenseqa_evaluation} reports the results.
Both models maintain comparable performance after pre-training, with Llama 3.1 8B slightly improving (71.8 $\rightarrow$ 72.7) and Qwen2 7B showing a minor decrease (80.7 $\rightarrow$ 79.5).
These results demonstrate that pre-training on negated corpora does not lead to catastrophic forgetting of general commonsense knowledge.

\begin{table}[h]
  \centering
  \small
  \begin{tabular}{l r}
\toprule
                                  & CommonsenseQA \\
\midrule
Llama 3.1 8B                     & 71.8 \\
~~~Pre-trained with \\
~~~~~~$\neg$\textsc{Atomic} + $\neg$\textsc{Anion}
                                  & 72.7 \\
                                  \addlinespace
Qwen2 7B                         & 80.7 \\
~~~Pre-trained with \\
~~~~~~$\neg$\textsc{Atomic} + $\neg$\textsc{Anion}
                                  & 79.5 \\
\bottomrule
\end{tabular}

  \caption{Results on CommonsenseQA (accuracy), a non-negated commonsense reasoning benchmark. Pre-training on our negated corpora does not degrade performance.}
  \label{t:commonsenseqa_evaluation}
\end{table}

\subsection{Ablation on Randomly Labelled Data}
\label{app:random_label}

Although our LLM judge achieves relatively high precision on validating commonsense triples with negation, the resulting data are still noisy.
To quantify the impact of data quality, we train models on a noisier dataset---with randomly assigned labels---where the validation accuracy for \textit{Valid} and \textit{Invalid} triples is approximately 0.50.

Table~\ref{t:random_label_ablation_study} reports the results.
Training with randomly labelled data substantially degrades performance compared to LLM-validated data across all tasks.
Most notably, NevIR drops well below the off-the-shelf baseline for both models (18.9 vs.\ 30.6 for Llama 3.1 8B; 12.0 vs.\ 29.8 for Qwen2 7B),
indicating that noisy labels can be actively harmful for complex negation reasoning.
For NLI tasks, randomly labelled data still yields some improvement over the baseline (e.g., RTE: 70.8 vs.\ 60.0 for Llama 3.1 8B),
suggesting that the model learns partial negation patterns from the triple structure itself.
However, the gains are far smaller than with validated data (e.g., RTE: 70.8 vs.\ 81.3).
These results validate the importance of our LLM judge: even imperfect labels substantially outperform random ones.

\begin{table*}[t]
  \centering
  \small
  \begin{tabular}{l r r r r r}
\toprule
                                  & \multirow{2}{*}{CondaQA}  & \multicolumn{3}{c}{NLI with Negation}       & \multirow{3}{*}{NevIR} \\
  \cmidrule(lr){3-5}
                                  & (Accuracy)  &                              RTE-Neg  &         SNLI-Neg &         MNLI-Neg                 \\
\midrule
Llama 3.1 8B                     &        68.7 &            60.0 &            54.4  &         47.0     &         30.6 \\
~~~Pre-trained with              \\
~~~~~~Randomly labelled $\neg$\textsc{Atomic} + $\neg$\textsc{Anion}
                                  &        67.1 &            70.8 &            57.5  &         50.9     &         18.9 \\
~~~~~~$\neg$\textsc{Atomic} + $\neg$\textsc{Anion}
                                  &        71.5 &            81.3 &            67.9  &         63.7     &         38.2 \\
                                  \addlinespace

Qwen2 7B                         &        65.1 &            71.7 &            60.7  &         59.5     &         29.8 \\
~~~Pre-trained with              \\
~~~~~~Randomly labelled $\neg$\textsc{Atomic} + $\neg$\textsc{Anion}
                                  &        65.2 &            76.1 &            62.0  &         61.9     &         12.0 \\

~~~~~~$\neg$\textsc{Atomic} + $\neg$\textsc{Anion}
                                  &        66.6 &            82.1 &            72.2  &         67.5     &         38.5 \\

\bottomrule
\end{tabular}

  \caption{
      Ablation on randomly labelled training data. We train Llama 3.1 8B and Qwen2 7B with randomly labelled $\neg$\textsc{Atomic} + $\neg$\textsc{Anion}.
      While the randomly labelled dataset demonstrates marginal benefits on CondaQA and NLI tasks, training with the validated dataset significantly outperforms.
      In addition, randomly labelled data is detrimental to NevIR performance.
}
  \label{t:random_label_ablation_study}
\end{table*}

\section{Error Analysis}
\label{app:error_analysis}

We conduct a detailed error analysis comparing the off-the-shelf Llama 3.1 8B with the model pre-trained on $\neg$\textsc{Atomic} + $\neg$\textsc{Anion} across all downstream tasks.
Our analysis categorizes the types of negation reasoning errors that are corrected by training with our augmented corpora.

\subsection{Error Types in NLI with Negation}
\label{app:error_types_nli}

We examine all 940 examples across RTE-Neg, SNLI-Neg, and MNLI-Neg where the pre-trained model answers correctly but the off-the-shelf model does not.
Following the negation taxonomy of~\citet{hossain-etal-2020-analysis}, we categorize these improvements by negation type in Table~\ref{t:error_analysis}.

\begin{table}[t]
  \centering
  \small
  \resizebox{0.48\textwidth}{!}{%
\begin{tabular}{l r r}
\toprule
Error Type & Count & Percentage \\
\midrule
Verbal negation in premise only    & 323 & 34.4\% \\
Verbal negation in hypothesis only & 220 & 23.4\% \\
Negation interaction (both P \& H) & 388 & 41.3\% \\
Affixal negation                   &   2 &  0.2\% \\
Other                              &   7 &  0.7\% \\
\bottomrule
\end{tabular}
}

  \caption{Distribution of error types corrected by pre-training with our commonsense corpora, across 940 improved NLI examples (Llama 3.1 8B). Categories follow the negation taxonomy of~\citet{hossain-etal-2020-analysis}.}
  \label{t:error_analysis}
\end{table}

\paragraph{Verbal Negation in Premise Only (34.4\%)}
In these cases, the premise contains an explicit verbal negation cue (e.g., \textit{not}, \textit{never}, \textit{no}) but the hypothesis does not.
The off-the-shelf model often treats the negated premise as if the negation were absent.
For example, given the premise \textit{``The prosecutor did not tell the court that the incident had caused `distress' to one of the children.''}
and the hypothesis \textit{``The prosecutor told the court that `distress' in one of the children is associated with the incident.''},
the off-the-shelf model predicts \textit{entailment}, ignoring \textit{``did not''} in the premise.
The pre-trained model correctly predicts \textit{not\_entailment}.

\paragraph{Verbal Negation in Hypothesis Only (23.4\%)}
These examples contain verbal negation in the hypothesis but not in the premise.
The off-the-shelf model struggles to determine whether a non-negated premise supports or contradicts a negated hypothesis.

\paragraph{Negation Interaction (41.3\%)}
The largest category involves examples where both the premise and hypothesis contain negation, requiring the model to reason about the interaction between multiple negation cues.
For instance, given \textit{``This growing number of titles does not leave publishing houses with less time and attention to edit and market books.''}
and \textit{``Publishing houses cannot give less attention to editing books.''},
the off-the-shelf model predicts \textit{entailment}.
However, the correct label is \textit{neutral}, as the negated premise no longer supports the negated hypothesis.
The pre-trained model correctly identifies this relationship.

\paragraph{Affixal Negation and Other (0.9\%)}
A small fraction of improvements involve affixal negation (e.g., \textit{un-}, \textit{dis-}) or other negation types.
This low proportion reflects the dominance of verbal negation cues in NLI benchmarks~\cite{hossain-etal-2020-analysis}.

\subsection{Analysis by Negation Cue Type in CondaQA}
\label{app:condaqa_cue_analysis}

CondaQA annotates each example with a negation cue word.
Following the negation taxonomy of~\citet{hossain-etal-2020-analysis}, we categorize these cues into five linguistic types and report accuracy for the off-the-shelf and pre-trained Llama 3.1 8B in Table~\ref{t:condaqa_negation_cue_analysis}.

\begin{table*}[t]
  \centering
  \small
  \begin{tabular}{l l r r r}
\toprule
Category & Examples & Count & Off-the-shelf & Pre-trained \\
\midrule
Verbal     & \textit{not, never, no, none}           & 1,902 & 67.6 & 74.1 \\
Affixal    & \textit{un-, in-, dis-, -less}           & 3,553 & 62.1 & 68.0 \\
Implicit   & \textit{lack, without, prevent}          & 1,502 & 67.9 & 69.7 \\
Diminisher & \textit{rarely, barely, few}             &   259 & 66.0 & 68.3 \\
Other      &                                          &    24 & 79.2 & 75.0 \\
\bottomrule
\end{tabular}

  \caption{CondaQA accuracy by negation cue category (Llama 3.1 8B). Categories follow the negation taxonomy of~\citet{hossain-etal-2020-analysis}: \textit{Verbal}: explicit negation words; \textit{Affixal}: morphological negation (prefixes/suffixes); \textit{Implicit}: words conveying negation without explicit markers; \textit{Diminisher}: words that reduce the degree of an assertion.}
  \label{t:condaqa_negation_cue_analysis}
\end{table*}

Pre-training with our corpora yields the largest improvement for verbal negation cues (+6.5 points), which aligns with our training data where negation is introduced by adding \textit{not} to events.
Affixal negation also shows substantial gains (+5.9 points), indicating effective transfer from explicit to morphological negation.
Diminisher cues (e.g., \textit{rarely}, \textit{barely}, \textit{few}) improve by +2.3 points, suggesting that learning explicit negation patterns helps models better handle attenuated assertions.
Implicit negation cues show a modest improvement (+1.8 points), indicating that implicit negation such as \textit{lack} or \textit{prevent} partially benefits from explicit negation training but may require additional training signals.
The Other category (n=24) is too small for reliable estimates.

\subsection{Case Studies}
\label{app:case_studies}

We present representative case studies illustrating specific reasoning patterns improved by our approach.

\paragraph{Case 1: Negation Scope in NLI}
\begin{quote}
\small
\textbf{Premise:} \textit{``A barefoot young girl in a pink gown is \textbf{not} asleep on a hard wood floor cuddling her baby doll.''}\\
\textbf{Hypothesis:} \textit{``A girl is playing with her doll outside.''}\\
\textbf{Gold:} neutral \\ \textbf{Off-the-shelf:} contradiction \\ \textbf{Pre-trained:} neutral
\end{quote}
The off-the-shelf model interprets \textit{``not asleep''} as contradicting \textit{``playing outside''}, failing to recognize that negating \textit{asleep} does not specify what the girl is doing or where she is.
The pre-trained model correctly identifies that the negated premise leaves the hypothesis unresolved.

\paragraph{Case 2: Negation and Entailment Direction}
\begin{quote}
\small
\textbf{Premise:} \textit{``Organic fertilizers like vermi compost are \textbf{not} used for increasing the quality, fertility and mineral content of the soil.''}\\
\textbf{Hypothesis:} \textit{``Organic fertilizers are used as soil enhancers.''}\\
\textbf{Gold:} not\_entailment \\ \textbf{Off-the-shelf:} entailment \\ \textbf{Pre-trained:} not\_entailment
\end{quote}
The off-the-shelf model associates \textit{organic fertilizers} with \textit{soil enhancers} based on world knowledge, completely ignoring the negation in the premise.
Pre-training on commonsense triples with negation teaches the model that a negated \emph{if} event does not entail the original \emph{then} event.

\paragraph{Case 3: Affixal Negation in CondaQA}
\begin{quote}
\small
\textbf{Cue:} \textit{unmyelinated}\\
\textbf{Question:} \textit{``Does the wording of the passage suggest that axons have myelin sheaths while neurons typically do not?''}\\
\textbf{Gold:} No \\ \textbf{Off-the-shelf:} Yes \\ \textbf{Pre-trained:} No
\end{quote}
The prefix \textit{un-} in \textit{unmyelinated} reverses the meaning, and the pre-trained model correctly identifies that the passage does not support the hypothesis.

\paragraph{Case 4: Negation in Information Retrieval (NevIR)}
\begin{quote}
\small
\textbf{Q1:} \textit{``Whose ship is attacked by \textbf{unfamiliar} enemies?''}\\
\textbf{Q2:} \textit{``Whose ship is attacked by \textbf{familiar} enemies?''}\\
\textbf{Expected:} Q1$\rightarrow$Doc1, Q2$\rightarrow$Doc2
\end{quote}
NevIR requires the model to distinguish between a query and its negated counterpart.
The model must identify which document contains information matching the specific polarity of each query---a task that directly requires understanding whether an event is negated.

\end{document}